\documentclass[journal]{IEEEtran}
\ifCLASSINFOpdf
\else
\fi
\usepackage{cite}
\usepackage{amsmath,amssymb,amsfonts}
\usepackage{algorithmic}
\usepackage{graphicx}
\usepackage{subcaption}
\usepackage{textcomp}
\usepackage{wrapfig}
\usepackage{algorithm}
\usepackage{lipsum}
\usepackage[hidelinks]{hyperref}

\usepackage{xcolor}
\usepackage{tabularx,booktabs}
\newcolumntype{C}{>{\centering\arraybackslash}X} 
\def\BibTeX{{\rm B\kern-.05em{\sc i\kern-.025em b}\kern-.08em
    T\kern-.1667em\lower.7ex\hbox{E}\kern-.125emX}}
\begin{document}
%
\title{A Low-Complexity Radar Detector Outperforming OS-CFAR for Indoor Drone Obstacle Avoidance} 
%
%
%

\author{Ali Safa, \IEEEmembership{Student Member, IEEE}, 
Tim Verbelen, 
Lars Keuninckx, 
Ilja Ocket, 
\\
\IEEEmembership{Member, IEEE}, 
Matthias Hartmann, Andr\'e Bourdoux, \IEEEmembership{Senior Member, IEEE},
\\

Francky Catthoor, \IEEEmembership{Fellow, IEEE}, 
Georges G.E. Gielen, \IEEEmembership{Fellow, IEEE}

\thanks{Ali Safa, Ilja Ocket, Francky Catthoor and Georges G.E Gielen are with imec and the Department of Electrical Engineering, KU Leuven, 3001 Leuven, Belgium (e-mail: Ali.Safa@imec.be; Ilja.Ocket@imec.be; Francky.Catthoor@imec.be; Georges.Gielen@kuleuven.be).}
\thanks{Tim Verbelen is with imec and the Department of Electrical Engineering, U Gent, 9000 Gent, Belgium (e-mail: Tim.Verbelen@imec.be).}
\thanks{Lars Keuninckx, Matthias Hartmann and Andr\'e Bourdoux are with imec, 3001 Leuven, Belgium (e-mail: Lars.Keuninckx@imec.be; Matthias.Harttmann@imec.be; Andre.Bourdoux@imec.be).}}
%
%

\markboth{}%
{Safa \MakeLowercase{\textit{et al.}}: A Simple Radar Detection Method Outperforming OS-CFAR for Indoor Drone Navigation}
%



\maketitle

\begin{abstract}
As radar sensors are being miniaturized, there is a growing interest for using them in indoor sensing applications such as indoor drone obstacle avoidance. In those novel scenarios, radars must perform well in dense scenes with a large number of neighboring scatterers. Central to radar performance is the detection algorithm used to separate targets from the background noise and clutter. Traditionally, most radar systems use conventional CFAR detectors but their performance degrades in indoor scenarios with many reflectors. Inspired by the advances in non-linear target detection, we propose a novel high-performance, yet low-complexity target detector and we experimentally validate our algorithm on a dataset acquired using a radar mounted on a drone. We experimentally show that our proposed algorithm drastically outperforms OS-CFAR (standard detector used in automotive systems) for our specific task of indoor drone navigation with more than 19\% higher probability of detection for a given probability of false alarm. We also benchmark our proposed detector against a number of recently proposed multi-target CFAR detectors and show an improvement of 16\% in probability of detection compared to CHA-CFAR, with even larger improvements compared to both OR-CFAR and TS-LNCFAR in our particular indoor scenario. To the best of our knowledge, this work improves the state of the art for high-performance yet low-complexity radar detection in critical indoor sensing applications.
\end{abstract}


\section*{Supplementary Material}
\noindent
An evaluation code 
and videos 
showcasing hardware metrics and our system implementation will be available at:
\\
\url{https://github.com/ali20480/RadarDetector}

%
\IEEEpeerreviewmaketitle

\section{Introduction}
\label{sec:introduction}
\IEEEPARstart{T}{he} use of radar sensors for indoor applications has been enabled recently thanks to the enormous progress in radar miniaturization \cite{c1,c2} and energy efficiency \cite{c3}, making use of the \textit{Frequency Modulated Continuous Wave} (FMCW) design principle \cite{c4}. Among those indoor sensing applications, the navigation of small \textit{Unmanned Aerial Vehicles} (UAVs) or drones is extensively being investigated for applications ranging from automated logistics \cite{b1} and indoor maintenance inspection \cite{b2} to people search and rescue in partially damaged infrastructures \cite{b3} all requiring robust obstacle avoidance. Currently, most autonomous drone systems solely use camera-based sensing to perform obstacle avoidance, path planning \cite{b4} and \textit{Simultaneous Localization and Mapping} (SLAM) \cite{b7}. As standard cameras have sensory limitations such as sensitivity to lighting conditions, occlusion by dirt and no intrinsic ranging \cite{b9}, many teams are investigating the use of complementary sensors such as \textit{Light Detection and Ranging} (LIDARs) \cite{b10}, Dynamic Vision Sensors (DVS) \cite{surv} and radars \cite{b18}. The latter is the focus of this work. Radars are well-suited as a complementary sensing modality because, unlike cameras, they intrinsically provide range information, are very robust towards environmental effects and are independent of lighting conditions \cite{throughwall}. Usually, radar sensors are tightly coupled to pre-processing units that provide range, velocity and angle of arrival spectra (so-called range-Doppler-angle profiles), which must be filtered afterwards by detection algorithms in order to separate target peaks from the background noise and clutter \cite{book}. 

Traditional detectors used within the radar community are the so-called \textit{Constant False Alarm Rate} (CFAR) class of algorithms, from low-complexity \textit{Cell Averaging} (CA) CFAR to more computationally complex \textit{Ordered Statistics} (OS) CFAR \cite{book}. As CA-based methods require a minimal separation distance between targets, OS-CFAR has been proposed as a better approach to the CA-CFAR based methods for multi-target scenarios where reflecting objects are closer to each other \cite{oscfa}. OS-CFAR has thus widely been adopted in critical areas such as \textit{outdoor} automotive applications as its probability of detection $P_{D}$ for a given probability of false alarm $P_{FA}$ is larger compared to CA-based methods \cite{oscar}. On the other hand, \textit{indoor} scenes are usually composed of a much larger number of reflecting objects placed closer to each other compared to the outdoor case. During our experiments, we have observed that OS-CFAR fails to detect all potential targets systematically when processing radar signals during an indoor drone flight (a discussion is provided in section \ref{discusssec}), while achieving a $P_D$ close to $1$ is essential for a safe obstacle avoidance. Thus, reliable radar target detection remains a challenge as missed detections can jeopardize the obstacle avoidance mechanisms and may crash the drone. Therefore, there is a need for a better, yet low-complexity detection algorithm featuring a higher $P_D$ for a certain $P_{FA}$ compared to the conventional CFARs, not only including OS-CFAR but also more recent, state-of-the-art CFAR variants such as OR-CFAR \cite{orcfar}, TS-LNCFAR \cite{lncfar} and CHA-CFAR \cite{chacfar}. 

Finally, it is important to clarify that our aim is to achieve a safe \textit{obstacle avoidance} rather than \textit{odometry}. When performing \textit{odometry} (i.e., indoor positioning), it is well known that detecting and tracking all potential targets (even the weak ones) may worsen the positioning results (only good, repeatable features must be tracked). On the other hand, when performing \textit{obstacle avoidance} like in our case, detecting all potential obstacles ($P_D$ close to $1$) is key for a safe drone flight (false alarms can be further removed by fusion with other sensing modalities). 

This paper is organized as follows. Section \ref{sec0} briefly reviews the related state-of-the-art works. Section \ref{sec1} describes the background theory. Section \ref{sec2} presents our proposed methods. Section \ref{sec3} describes our experimental validation. Finally, section \ref{sec4} concludes this article.

\section{Related Works}
\label{sec0}
To the best of our knowledge, radar sensing for drones has mainly been investigated for \textit{outdoor} navigation above wasteland containing a sparse number of ground objects (while our work addresses the indoor setting). Most recent among those works, the authors in \cite{b18} demonstrated outdoor drone navigation using an \textit{Inertial Measurement Unit} (IMU, magnetometer, gyroscope and accelerometer), a camera and a 24-GHz FMCW radar, and reported a lower RMS navigation error against the ground-truth GPS compared to older radar-based drone solutions \cite{b16,b17}. In addition, the authors in \cite{b18} clearly demonstrated that conventional CFAR filters can suffer from a large number of missed detections and therefore proposed an alternative detection mechanism by first filtering the radar maps with a Gaussian kernel and then keeping the \textit{five largest peaks} as detection. While such detection algorithm is well suited for navigation above wasteland containing an \textit{a priori} known number of targets, it is much less the case for the dense indoor setting where the (potentially large) number of scatterers is unknown. 

Regarding the use of radar in indoor drone navigation, the authors in \cite{SARR} proposed an \textit{indoor} FMCW SAR imaging system mounted on a drone and tested in an indoor setting with four corner reflectors and one large obstacle placed in the middle of the scene. Although promising, a challenge faced with such system is the lack of sufficient space in constrained indoor environments in order to form a large aperture. In addition, as only a few strong reflectors were present in the scene, it remains unclear how the system would behave in realistic indoor scenarios in terms of effective detection and false alarms as no explicit detection is performed (only imaging is provided). Another work aiming at indoor SAR imaging using FMCW radars was proposed in \cite{SARR2}, still without a discussion on radar target detection. 

As conventional CFAR filters can severely suffer from missed detection, a large number of alternative methods have been proposed. The authors in \cite{comb} proposed a detector which combines multiple conventional CFAR variants and chooses the best one depending on the clutter homogeneity. The authors in \cite{regress} proposed to estimate the detection threshold of conventional CFARs using polynomial regression on the maps to be filtered, which, compared to the conventional CA-CFAR, provided a better threshold estimate as higher statistical moments are used during the estimation. 

Recently, a number of CFAR detectors based on the \textit{truncated statistics} principle have been proposed for multi-target detection in outdoor environments (e.g., ship detection from SAR images), using a careful modelling of the clutter statistics \cite{orcfar, lncfar,chacfar}. In addition to OS-CFAR, the applicability of those aforementioned methods to indoor scenarios is also investigated in this work. Using modern machine learning techniques, a CFAR detector using Support Vector Machines (SVMs) was proposed in \cite{scm} and a neural network-based CFAR was proposed in \cite{nncfar}. Although highly efficient in scenarios well captured by the training data, those methods based on supervised training are not guaranteed to perform well across radically different scenes. 

Closely related to our work, the authors in \cite{kerrx} proposed the \textit{kernel RX} (KRX) detector as an enhancement of the classical \textit{Reed-Xiaoli} (RX) detector used in hyperspectral imagery \cite{rx}. They showed that KRX significantly outperforms RX (in terms of the reachable $P_D$ for a certain $P_{FA}$) without being significantly more computationally complex. Still, the KRX detector can be too computationally expensive, as it requires a large number of matrix inversions \cite{kerrx}. Therefore, we have derived our detector based on the same principles used within KRX while proposing several modifications to simplify the computational cost of the algorithm, such that implementation is possible in extreme edge devices, in contrast to the KRX detector. 
\section{Background Theory}
\label{sec1}
First, the FMCW radar theory and pre-processing are briefly introduced. Then, the conventional CFAR detectors are briefly discussed. Finally, the principles behind the KRX detector are presented. These are needed to explain our method in the next section.
\subsection{FMCW Radar}
In Single Input Multiple Outputs (SIMO) mode, a radar uses one TX antenna and $m=1,...,N_{RX}$ RX antennas, usually spaced by an inter-antenna distance $d_a=\frac{\lambda}{2}$ where $\lambda$ is the radar carrier wavelength \cite{antenna}. In each sensing frame, the radar sends a sequence of $q = 1,...,N_c$ chirps defined by:
\begin{equation}
    p_q(t) = \exp{j(2 \pi f_c t + \pi \alpha t^2)}
    \label{chirp}
\end{equation}
where $f_c$ is the radar carrier frequency (77-GHz in this work) and $\alpha$ is the chirp slope \cite{exa2}. By reflection on surrounding objects, a mix of delayed and attenuated versions of $p_q(t)$ is received at each RX antennas. Those received signals are demodulated by the IQ receiver such that for each chirp $q$ and each antenna $m$, an \textit{Intermediate Frequency} (IF) signal is obtained as follows:
\begin{equation}
    r_{qm}(t) = \sum_{i=1}^{N_{t}} \xi_i e^{j(-2 \pi \alpha T_{d_i}  t - 2 \pi f_c T_{d_i}  + \pi \alpha T_{d_i}^2 + m\phi_i)}
    \label{eq2}
\end{equation}
where $N_t$ is the number of target reflectors in the scene, $\xi_i$ is the attenuation of the target reflector $i$, $\phi_i$ is the phase shift of target $i$ induced by its angle of arrival, and $T_{d_i}$ is the round-trip time from the radar antenna to the target $i$, which is proportional to the distance $d_i$ between the radar and the target as $T_{d_i} = \frac{2d_i}{c}$ (where $c$ is the speed of light) \cite{book}. The phase shift $\phi_i$ for target $i$ is given by $\phi_i = \pi \sin{\theta_i}$ \cite{antenna} where $\theta_i$ is the angle of arrival of the signal reflected by target $i$. The IF signals $r_{qm}(t)$ (\ref{eq2}) are then sampled by an ADC with a sampling period $T_f$ to obtain the discrete-time signals $r_{qm}[n]$ for each chirp $q$ and each antenna $m$. It is then possible to retrieve the range, Doppler radial velocity and angle of arrival of each target $i$ by computing the Discrete Fourier Transform (DFT) of $r_{qm}[n]$ along $n$, $q$ and $m$, respectively (a 3D DFT for each radar frame) \cite{exa3}. The magnitude of the resulting data cube exhibits $N_t$ peaks at the bin locations corresponding to the range, velocity and angle of arrival of each target $i$ in the scene. 

Noise, clutter (ground echo) and multipath will also result in additional undesired peaks that must be filtered using a detection algorithm such as conventional CFAR algorithms \cite{book}. To reduce the number of DFT and detection operations (and thus, to increase the maximal frame rate achievable by the sensing system), the 3D DFT is not done at once; rather, a 1D DFT is first performed along $n$ to obtain the so called \textit{range profile} (RP) with its magnitude containing peaks at target locations. The magnitude of the RP is then filtered by a detector such as a CFAR, obtaining a subset of \textit{populated} range bins. It is then possible to perform the DFTs along $q$ and $m$ for each populated range bin only, to reduce the computational costs. It must be noted that this solution is well-suited when the relative velocity between the radar and the obstacles is relatively low so that the target remains within the same range bin at different time epochs, which is the case in indoor navigation, 
where the drone speed is relatively low due to the size of the drone and the weight of the payloads (for a drone flying at $10$kph with a typical coherent processing interval of $12$ms as used in our radar setup, the maximum range migration of obstacles will be about $2.8$m/s$\times 0.012=3.36$cm which falls below the range resolution of $7.1$cm used in our radar setup). 

Another, more expensive approach is to first perform the 2D DFT along $n$ and $q$ to obtain the so-called \textit{range-Doppler profile} (RDP). The RDP is then filtered along each range and Doppler bin to find the subset of populated bins. As this second method introduces an additional decorrelation, conventional CFAR detectors will perform better at the expense of a severe increase in computation (i.e., the detection in this second approach must be performed along each range and Doppler bin with typical dimensions of e.g. $256 \times 128$ bins versus only one detection pass for the first approach) \cite{exp4}. 

As high frame rates are desired for our indoor drone application to increase the dynamic responsiveness of the drone, we therefore adopt the first approach even though this method trades off lower signal processing gain for higher speed. In addition, this will allow us to assess the performance of our algorithms on more challenging data in section \ref{sec3} as a second decorrelation introducing the additional Doppler processing gain is not introduced.

\subsection{Conventional CFAR detectors}
\label{secchal}
\begin{figure}[htbp]
\centerline{\includegraphics[scale=0.6]{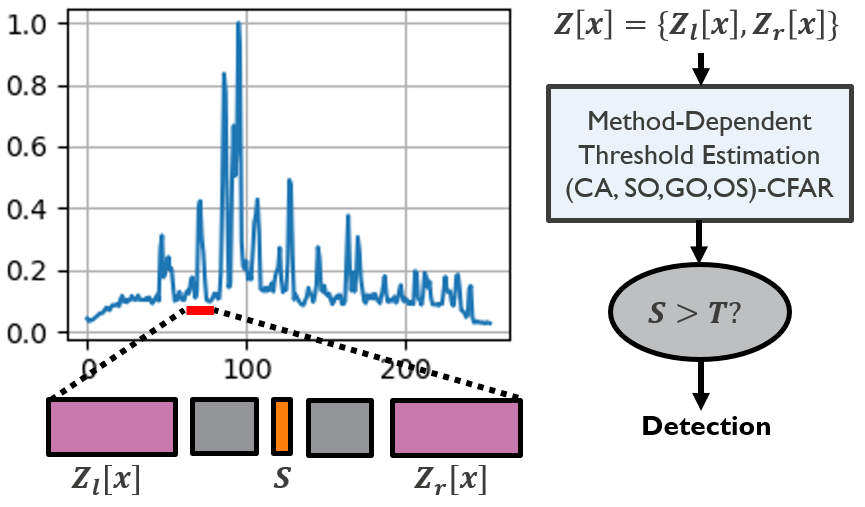}}
\caption{\textbf{A typical range profile being filtered by a CFAR detector}. The CUT is the cell $S$ in the middle, the training cells form a vector $Z[x]$ and the remaining bins are the guard cells.}
\label{figtree}
\end{figure}

CFAR filters are used to detect targets in unknown background noise \cite{book}. The most traditional CFAR is the CA-CFAR, which computes a local detection threshold for the \textit{cell under test} (CUT) proportional to the average of the local \textit{training} cells in a sliding analysis window (see Fig. \ref{figtree} where $S$ is the CUT and the $Z_l[x]$ and $Z_r[x]$ are the left and right training cells, respectively).
For locally open environments, where scatterers are well isolated, CA-CFAR performs well, but its performance can strongly degrade in non-homogeneous environments where multiple targets and clutter edges are present \cite{scm}. The \textit{greatest of} (GO) CA-CFAR (which chooses the greatest average between the left training cells $Z_l[x]$ and the right training cells $Z_r[x]$) performs better compared to CA-CFAR in environments with several clutter edges but suffers from a loss in $P_D$ when mulitple targets are present locally. On the other hand, \textit{smallest of} (SO) CA-CFAR (which chooses the smallest average between the left training cells $Z_l[x]$ and the right training cells $Z_r[x]$) performs better in the multi-target scenario but is very sensitive to clutter edges, resulting in a significant increase in $P_{FA}$ \cite{book}. 

It has been reported that the OS-CFAR performs significantly better than CA-based CFARs for the multi-target case (its $P_D$ is larger than CA-CFAR for a certain $P_{FA}$) and therefore has widely been adopted in (outdoor) automotive applications \cite{oscfa,oscar,carcomp}. OS-CFAR first sorts all the training cells in increasing order and selects the $k^{th}$ element as the noise estimate $\hat{\beta}^2$. Then, the local detection threshold is derived as $T=\alpha \hat{\beta}^2$ where $\alpha$ sets the $P_{FA}$ and $k$ is a parameter to be chosen. 

Clearly, as this method requires a sorting operation per CUT, it is significantly more computationally expensive than CA-based methods \cite{carcomp}. OS-CFAR has a typical complexity of $ O(L N_{tc}^2)$ when using \textit{Bubble sorting} and $ O(L N_{tc} \ln{(N_{tc})})$ when using \textit{Quick sorting} where $N_{tc}$ is the number of training cells (noted as $Z[x]$ in Fig. \ref{figtree}) and $L$ is the total number of range bins \cite{oscfa}. On the other hand, CA-based methods have a complexity of $O(L)$ as \textit{multiply-accumulate} operations can be done in one clock cycle on CPUs with vectored instructions, unlike sorting. The goal of our proposed detector is to improve the performance beyond OS-CFAR, while reducing the computational complexity.
In section \ref{sec3}, we will benchmark our proposed detector against OS-CFAR as it is the standard detector adopted in most multi-target applications \cite{book}.

\subsection{State-of-the-art CFAR principles for multi-target scenarios}
\label{sotalol}
Even though still widely used in practice, the conventional CFAR detectors discussed in section \ref{secchal} have shortcomings in a number of situations where their clutter statistics estimation performance becomes poor (multi-target environments, non-homogeneous clutter and so on) \cite{itshard}. Thanks to the enormous progress made by the remote sensing community, a number of novel radar detectors specially tailored for the multi-target scenario have been proposed, achieving state-of-the-art result in anomaly detection from satellite SAR images \cite{orcfar,lncfar,chacfar,gammacfar}. Among those, we consider three candidate detectors in this work, in addition to OS-CFAR. 
\subsubsection{Outlier-robust CFAR (OR-CFAR)}
OR-CFAR \cite{orcfar} has been proposed as a well-suited detector for \textit{Gaussian clutter} scenarios subject to an important number of outliers that could jeopardize the noise estimation performance of conventional CFAR methods. OR-CFAR achieves a robust noise estimation performance by using a truncated statistics (TS) mechanism where outliers in the training cells $Z[x]$ are removed if:
\begin{equation}
    Z[x] - \mu > \gamma \sigma 
    \label{eqtrunc}
\end{equation}
where $\gamma$ is the truncation depth, $\mu$ is the mean of the training cells and $\sigma$ their standard deviation. Then, the \textit{maximum-likelihood} (ML) estimate of the clutter mean $\hat{\mu}$ and standard deviation $\hat{\sigma}$ from the remaining training cells $\Tilde{Z}[x]$ after truncation by (\ref{eqtrunc}) are given by \cite{orcfar}:

\begin{equation}
    \hat{\sigma} = \sqrt{\frac{1}{\chi}\{\frac{1}{n}\sum_{i=1}^{n}\Tilde{Z}[i]^2 - (\frac{1}{n}\sum_{i=1}^{n}\Tilde{Z}[i])^2\}}
\end{equation}

\begin{equation}
    \hat{\mu} = \frac{1}{n}\sum_{i=1}^{n}\Tilde{Z}[i] + \alpha \hat{\sigma}
\end{equation}

\begin{equation}
    \alpha = \frac{e^{-\frac{\gamma^2}{2}}}{\sqrt{2 \pi} \phi(\gamma)}, \hspace{10pt} \beta = 1-\gamma \alpha , \hspace{10pt} \chi = \frac{1}{\beta - \alpha^2}
\end{equation}
where $\phi(.)$ is the CDF of the standard normal distribution and $n$ the number of training cells after truncation. A CUT of amplitude $S$ is reported as a detection if $S>\hat{\mu} + t\hat{\sigma}$ where $t$ is a parameter setting the desired $P_{FA}$.
\\

\subsubsection{Truncated statistics log-normal CFAR (TS-LNCFAR)} 
TS-LNCFAR \cite{lncfar} has been proposed as a well-suited detector for \textit{log-normal clutter} scenarios subject to an important number of outliers. Similarly to OR-CFAR, TS-LNCFAR uses a TS approach for outlier rejection, followed by a ML estimation of the clutter mean and standard deviation. Outliers in the training cells are removed if:
\begin{equation}
    \ln Z[x] - \mu > \gamma \sigma 
    \label{eqtrunc}
\end{equation}
with $\gamma$ the truncation depth and $\mu$, $\sigma$ the mean and standard deviation of the log training cells $\ln Z[x]$. The ML estimate of the clutter mean $\hat{\mu}_{ln}$ and standard deviation $\hat{\sigma}_{ln}$ are then given 
by a set of equation similar to those in OR-CFAR \cite{lncfar}. A CUT of amplitude $S$ is reported as a detection if $\ln S>\hat{\mu}_{ln} + t\hat{\sigma}_{ln}$ where $t$ is a parameter setting the desired $P_{FA}$.  
\\



\subsubsection{Censored harmonic averaging CFAR (CHA-CFAR)}
CHA-CFAR \cite{chacfar} has been proposed as a well-suited detector for \textit{exponential clutter} scenarios affected by a large number of outliers. In contrast to methods that seek to \textit{hardly} remove outliers by truncation (such as in OR-CFAR and TS-LNCFAR), CHA-CFAR \textit{softly} removes the effect of outliers using the harmonic mean and the OS principle to estimate the noise level $\hat{\omega}$:

\begin{equation}
    \hat{\omega} = (z_{m+1}^{-1} + z_{m+2}^{-1} + ... + z_{N}^{-1})^{-1}
\end{equation}
where $\{z_1,...z_{N}\}$ is a set obtained by sorting the elements of $Z[x]$ in increasing order and $m$ is the number of discarded samples (with the smallest amplitude) \cite{chacfar}. A CUT of amplitude $S$ is reported as a detection if $S > t\hat{\omega}$ where $t$ is a parameter setting the desired $P_{FA}$.  
\\

As those state-of-the-art, multi-target techniques have mostly been demonstrated for earth and ocean observation, it is interesting to study their applicability in indoor environments. Therefore, in addition to OS-CFAR (standard detector for automotive applications), we will also benchmark our proposed detector against OR-CFAR, TS-LNCFAR and CHA-CFAR in section \ref{sec3}.     

\subsection{KRX detection principle}
\label{secrx}
The novel detector that we propose in this work (in section \ref{sec2}) is based on the principles behind \textit{kernel} RX (KRX) detection \cite{kerrx}. The KRX detector has initially been proposed for hyperspectral images, where each pixel $\Bar{x}$ is a vector with each vector entry corresponding to a certain spectral band (as opposed to the RP that we aim to filter in this work, obtained after applying range processing on (\ref{eq2}), where each pixel $\Bar{x}$ of the RP reduces to a scalar). In order to introduce KRX, the original RX detector, on which KRX is built, must be introduced first. The RX detector can be seen as a generalization of the CA-CFAR method to the multivariate Gaussian case (pixels are not scalars but rather vectors) \cite{rx}:
\begin{equation}
    RX(\Bar{x}) = (\Bar{x} - \hat{\Bar{\mu}})^T \hat{C}^{-1} (\Bar{x} - \hat{\Bar{\mu}})
    \label{rx}
\end{equation}
where a pixel $\Bar{x}$ is detected when $RX(\Bar{x}) > T$, where $T$ is a threshold setting the $P_{FA}$. $\hat{\Bar{\mu}}$ and $\hat{C}$ respectively denote the mean vector and the covariance matrix of the multivariate Gaussian noise estimated from the training cells in the sliding window (similar to Fig. \ref{figtree}). When the background statistics cannot be modelled by a single Gaussian distribution (as in our multi-target scenario), the RX detector performance degrades significantly \cite{nasr}. KRX was proposed in \cite{kerrx} as a nonlinear version of RX, which expresses the original RX model of (\ref{rx}) in a high-dimensional feature space $\mathcal{F}$ of dimension $D$ through a nonlinear transformation $\Phi(\Bar{x})$ as:
\begin{equation}
    KRX(\Bar{x}) = (\Phi(\Bar{x}) - \hat{\Bar{\mu}}_{\Phi})^T \hat{C}^{-1}_{\Phi} (\Phi(\Bar{x}) - \hat{\Bar{\mu}}_{\Phi})
    \label{krx}
\end{equation}
where $\hat{\Bar{\mu}}_{\Phi}$ and $\hat{C}_{\Phi}$ denote the estimated mean and covariance in $\mathcal{F}$. The covariance matrix in the feature space $\mathcal{F}$ is defined as $ \hat{C}_{\Phi} = (\Phi(X_{tc}) - \hat{\Bar{\mu}}_{\Phi})^T (\Phi(X_{tc})- \hat{\Bar{\mu}}_{\Phi})$ where $X_{tc}$ defines a matrix containing the training pixel vectors along each column and $\Phi(X_{tc})$ denotes the application of $\Phi$ to each of the individual vectors in $X_{tc}$. Eq. (\ref{krx}) is derived by considering that, once projected to $\mathcal{F}$, the model is well-approximated by a Gaussian distribution, even though the process is non-Gaussian in the original space \cite{nasr}. As $\mathcal{F}$ could possibly be of infinite dimensions, (\ref{krx}) is simplified using the \textit{kernel trick} \cite{abu}, which replaces the inner products found by expanding the terms in (\ref{krx}) by a \textit{kernel} function $k(\Bar{x}_i, \Bar{x}_j) = \langle \Phi(\Bar{x}_i),\Phi(\Bar{x}_j) \rangle $. 

A common choice for $k(\Bar{x}_i, \Bar{x}_j)$ is the \textit{Radial Basis Function} (RBF) $k(\Bar{x}_i, \Bar{x}_j) = \exp (-\gamma ||\Bar{x}_i - \Bar{x}_j||_2^2)$ for which the Gaussian assumption in $\mathcal{F}$ holds \cite{kerrx}. Thus, KRX nonlinearly projects the input data in a space $\mathcal{F}$ which is such that a simple threshold decision region in $\mathcal{F}$ corresponds to a much more complex decision region in the original space. This enables the algorithm to perform significantly better in environments with complex background statistics such as highly cluttered and multi-target indoor scenes \cite{kerrx}. Also, other hyperspectral detectors have been extended using the non-linear projection principle described above and it was demonstrated that the nonlinear versions always outperformed the original ones in terms of reaching a high $P_{D}$ for a given $P_{FA}$ \cite{compker}. 

Although promising for our near-sensor detection application, the approach given above still presents three challenges for implementation on resource-constrained computing platforms:
\begin{enumerate}
    \item Even when the kernel trick is used, (\ref{krx}) requires a large matrix inversion per CUT ($N_{tc} \times N_{tc}$) \cite{kerrx}, making is unsuited for fast near-sensor detection. 
    \item The RBF kernel is an exponential model requiring a high bit precision as its scale changes rapidly \cite{neuro}. Can we find a nonlinear transformation $\Phi$ well-suited for heavy quantization, even down to 1-bit precision?
    \item If such $\Phi$ that is well-suited for 1-bit quantization exists, the Gaussian assumption in the feature space will not hold anymore. How can we then correct for this initial assumption? 
\end{enumerate}
These challenges will be addressed by our proposed detector (see next section).
\section{Proposed Detector}
\label{sec2}
In this section, we derive our novel detector starting from (\ref{krx}). We first remark that (\ref{krx}) is the square of the Mahalanobis distance between $\Phi(\Bar{x})$ and $\hat{\Bar{\mu}}_{\Phi}$, which is a measure of \textit{distance} between $\Phi(\Bar{x})$ and the background statistics in $\mathcal{F}$ \cite{nasr}. When this distance is \textit{larger} than a threshold $T_0$, $\Bar{x}$ is reported as a detection. We can consider the dual problem \cite{review}: when the correlation between $\Phi(\Bar{x})$ and the background statistics in $\mathcal{F}$ is \textit{smaller} than a threshold $T_1$, $\Bar{x}$ is reported as a detection. The original detection problem:
\begin{equation}
    (\Phi(\Bar{x}) - \hat{\Bar{\mu}}_{\Phi})^T \hat{C}^{-1}_{\Phi} (\Phi(\Bar{x}) - \hat{\Bar{\mu}}_{\Phi}) > T_0
    \label{eqlog}
\end{equation}
then becomes:
\begin{equation}
    (\Phi(\Bar{x}) - \hat{\Bar{\mu}}_{\Phi})^T \hat{C}_{\Phi} (\Phi(\Bar{x}) - \hat{\Bar{\mu}}_{\Phi}) < T_1
    \label{eqrela}
\end{equation}
where the computationally expensive matrix inversion does not appear anymore. Eq. (\ref{eqlog}) and (\ref{eqrela}) are dual in the sense that searching for a $\Phi(\Bar{x})$ with a \textit{distance} \textit{larger} than a certain threshold from a distribution parametrized through $\hat{\Bar{\mu}}$ and $\hat{C}_{\Phi}$ (\ref{eqlog}) is conceptually the same as searching for a $\Phi(\Bar{x})$ with a \textit{correlation} \textit{smaller} than a certain threshold from the same distribution (\ref{eqrela}). It is important to note that, although solving the same problem, (\ref{eqlog}) and (\ref{eqrela}) are not identical \textit{per se} as they are solutions to different optimization problems 
 (see \cite{review} for a review). In order to simplify (\ref{eqrela}) even more, we choose a transformation $\Phi$ such that $\hat{\Bar{\mu}}_{\Phi} \xrightarrow{} \Bar{0}$ with $\hat{\Bar{\mu}}_{\Phi}$ defined as:
\begin{equation}
    \hat{\Bar{\mu}}_{\Phi} = \frac{1}{N_{tc}}\sum_{i=1}^{N_{tc}} \Phi(\Bar{x}_i)
    \label{eqdef}
\end{equation}
where $N_{tc}$ is the number of training cells in the window from which the noise estimation is performed. We therefore choose a $\Phi$ which induces \textit{sparse} projections $\Phi(\Bar{x})$ in $\mathcal{F}$, with only a few (possibly one) non-zero entries, such that $\hat{\Bar{\mu}}_{\Phi}$ can be dropped in (\ref{eqrela}). Under this \textit{sparse} projection, (\ref{eqrela}) becomes:
\begin{equation}
    \Phi(\Bar{x})^T \Phi(X_{tc}) \Phi(X_{tc})^T  \Phi(\Bar{x}) < T_1
    \label{eqlastt}
\end{equation}
where $X_{tc}$ is a $D \times N_{tc}$ matrix with $D$ the dimension of $\mathcal{F}$, containing the projection of the training cells in $\mathcal{F}$. We can further rewrite (\ref{eqlastt}) as:
\begin{equation}
    \Phi(\Bar{x})^T \Bar{N}_{\Phi} < T_1
    \label{eqlastt1}
\end{equation}
where $\Bar{N}_{\Phi}$ can be seen as a summary of the background noise statistics integrated through $\Phi(\Bar{x})$ against which the correlation is measured. As remarked in section \ref{secrx}, (\ref{eqlastt1}) holds under Gaussian assumption in the feature space. As we are aiming for a sparse $\Phi$, well-suited for 1-bit quantization, the Gaussian assumption does not hold and we need to modify how the background noise statistics are summarized in $\mathcal{F}$ through a modification of $\Bar{N}_{\Phi}$. 

To this end, we use the simple heuristic used by SOCA-CFAR, which states that the larger the norm of a pixel in the initial space, the less likely it is noise, such that its effect should be attenuated when summarizing the background noise statistics \cite{book}. Using this heuristic, we replace $\Bar{N}_{\Phi}$ by a \textit{centroid} across the training cells where each weight is inversely proportional to the $L_2$-norm of its corresponding training pixel:
\begin{equation}
   \Bar{N}_{\Phi}^m = \frac{1}{\Gamma} \sum_{j=1}^{N_{tc}} \frac{1}{||X_{tc}[j]||_2} \Phi(X_{tc}[j])
   \label{bad1}
\end{equation}
where $X_{tc}$ is assumed to be normalized to a maximum pixel $L_2$ norm of $1$, with
\begin{equation}
    \Gamma = \sum_{j=1}^{N_{tc}} \frac{1}{||X_{tc}[j]||_2}
     \label{bad2}
\end{equation}
being the amplitude normalization factor to effectively compute a \textit{centroid}. The test (\ref{eqlastt1}) then becomes:
\begin{equation}
    \Phi(\Bar{x})^T \Bar{N}_{\Phi}^m < T_1
     \label{bad3}
\end{equation}
Further simplifications can be made to reduce the amount of \textit{multiply} and \textit{divide} operations in (\ref{bad1}), (\ref{bad2}) and (\ref{bad3}). First, we can truncate the Taylor expansion of $x^{-1}$ around $1$ to the first-order term as $\frac{1}{x} \approx 2 - x $ where we can substitute $x=2||X_{tc}[j]||_2$ (with $x \in [0, 2]$ as we considered $||X_{tc}[j]||_2 \in [0,1]$) to obtain:
\begin{equation}
    \frac{1}{||X_{tc}[j]||_2} \approx 2(1 - ||X_{tc}[j]||_2) 
    \label{simpli}
\end{equation}
Therefore, instead of computing $\frac{1}{||X_{tc}[j]||_2}$, we will compute $(1 - ||X_{tc}[j]||_2)$ where the constant $2$ is dropped after normalization by $\Gamma$. Secondly, we can remove the expensive \textit{multiply-accumulate} operations in (\ref{bad3}) by solving the dual problem: a \textit{small} ($< T_1$) correlation between $\Phi(\Bar{x})$ and $\Bar{N}_{\Phi}^m$ corresponds to a large distance between $\Phi(\Bar{x})$ and $\Bar{N}_{\Phi}^m$:
\begin{equation}
    ||\Phi(\Bar{x}) - \Bar{N}_{\Phi}^m||_p > T_2
    \label{finfin}
\end{equation}

We choose $p=\infty$ instead of $p=2$ for two reasons. First, in order to replace the explicit norm computation involving \textit{multiply-accumulate} operations by a sequence of simple \textit{comparison} operations in parallel. Secondly, to attenuate the data-dependent bias introduced in the relation between the correlation in (\ref{bad3}) and the $L_2$ norm $||\Vec{a}-\Vec{b}||_2^2 = \sum_i a_i^2 + \sum_i b_i^2 - 2\sum_i \Vec{a}_i \Vec{b}_i$ where the term $\sum_i x_i y_i$ is the correlation and the term $\sum_i x_i^2 + \sum_i y_i^2$ is a data-dependent bias. The $L_\infty$-norm attenuates this bias as $||\Vec{a}-\Vec{b}||_\infty \leq ||\Vec{a}-\Vec{b}||_2$ always holds \cite{linalg}. At this point we have our final detector (\ref{finfin}). In the sliding window, the CUT $\Bar{x}$ will be reported as a detection if (\ref{finfin}) holds with $p=\infty$. 

Finally, we need to find a non-linear and sparse transformation $\Phi$. As our goal is to filter a radar RP in this work, the pixels $\Bar{x}$ that were generically considered as vectors become scalars that we note as $\xi = \Bar{x}$ in the remainder of the derivation. We found that a well-suited $\Phi$ is the one-hot encoding of $\xi$ as follows:
\begin{equation}
    \Phi(\xi)_i = \begin{cases} 1, & \mbox{if } i=D-\left \lfloor{\frac{D}{\xi}}\right \rfloor \\ 0, & \mbox{else}\end{cases}
    \label{finalphi}
\end{equation}
where $D$ is the dimension of the destination space $\mathcal{F}$. Such transformation is well-suited as it is binary (1-bit precision), sparse (as at most one element is non-zero, $\hat{\Bar{\mu}}_{\Phi}$ can be ignored in (\ref{eqrela})) and promotes linear independence between the pixel projections in $\mathcal{F}$ (which better complies with the constraint of non-singular covariance matrix $\hat{C}_{\Phi}$ in the original problem of KRX (\ref{eqlog})). 

In summary, our detection algorithm performs the following steps:
\begin{enumerate}
    \item $\Phi$ (\ref{finalphi}) is applied to the CUT and the training cells in the sliding window.
    \item $\Bar{N}_{\Phi}^m$ is computed following (\ref{bad1}), (\ref{bad2}) with the simplification of (\ref{simpli}).
    \item Each entry $i$ of the CUT projection $\Phi(\xi)[i]$ is compared to $T_2 + \Bar{N}_{\Phi}^m[i]$ and the CUT is reported as a detection as soon as an entry $i$ gives $\Phi(\xi)[i] > T_2 + \Bar{N}_{\Phi}^m[i]$ (by virtue of the infinite-norm in (\ref{finfin})).
\end{enumerate}

Similar to the original KRX detector, finding a closed-form relation between the threshold $T_2$ and the $P_{FA}$ is intractable, and the threshold must rather be selected empirically by 1) acquiring a labelled dataset 2) measuring the experimental $P_{FA}$ and $P_{D}$ values by sweeping the threshold and 3) selecting a threshold returning the desired $P_{FA}$ and $P_{D}$ \cite{nasr}.

The complexity of our proposed algorithm is $O(L D)$ compared to $ O(LN_{tc}^2)$ and $ O(LN_{tc} \ln{(N_{tc})})$ for OS-based methods (such as OS-CFAR and CHA-CFAR) and $ O(L)$ for CA-based methods (such as OR-CFAR and TS-LNCFAR) \cite{carcomp}, where $D$ is rather small as equation (\ref{finalphi}) projects scalars to vectors (we found $D \leq 15$ to be well-suited during our experiments, see section \ref{sec3}). 

Because of the simple heuristic rule used in (\ref{bad1}), we expect the performance of the algorithm to degrade for degenerated situations where an overwhelmingly large number of neighbouring targets are present, as most bins in the sliding window will then miss-represent the noise statistics (those situations are unlikely to happen through as realistic indoor scenes always have some empty space between obstacles). Nevertheless, we found this rule to be both necessary and efficient during our experiments in a regular indoor scenario: removing it was causing a severe performance degradation while incorporating it provided excellent detection results (see the \textit{ablation studies} in section \ref{abl}). 

\section{Experimental Performance Assessment}
\label{sec3}
In this section, we will experimentally assess the performance of our novel detector against OS-CFAR by generating the \textit{Receiver Operating Curves} (ROC: $P_D$ versus of $P_{FA}$) of both solutions on labelled radar data acquired using a drone flying in a dense indoor scene. 

\subsection{Experimental Setup}
\begin{figure}[htbp]
\centerline{\includegraphics[scale=0.13]{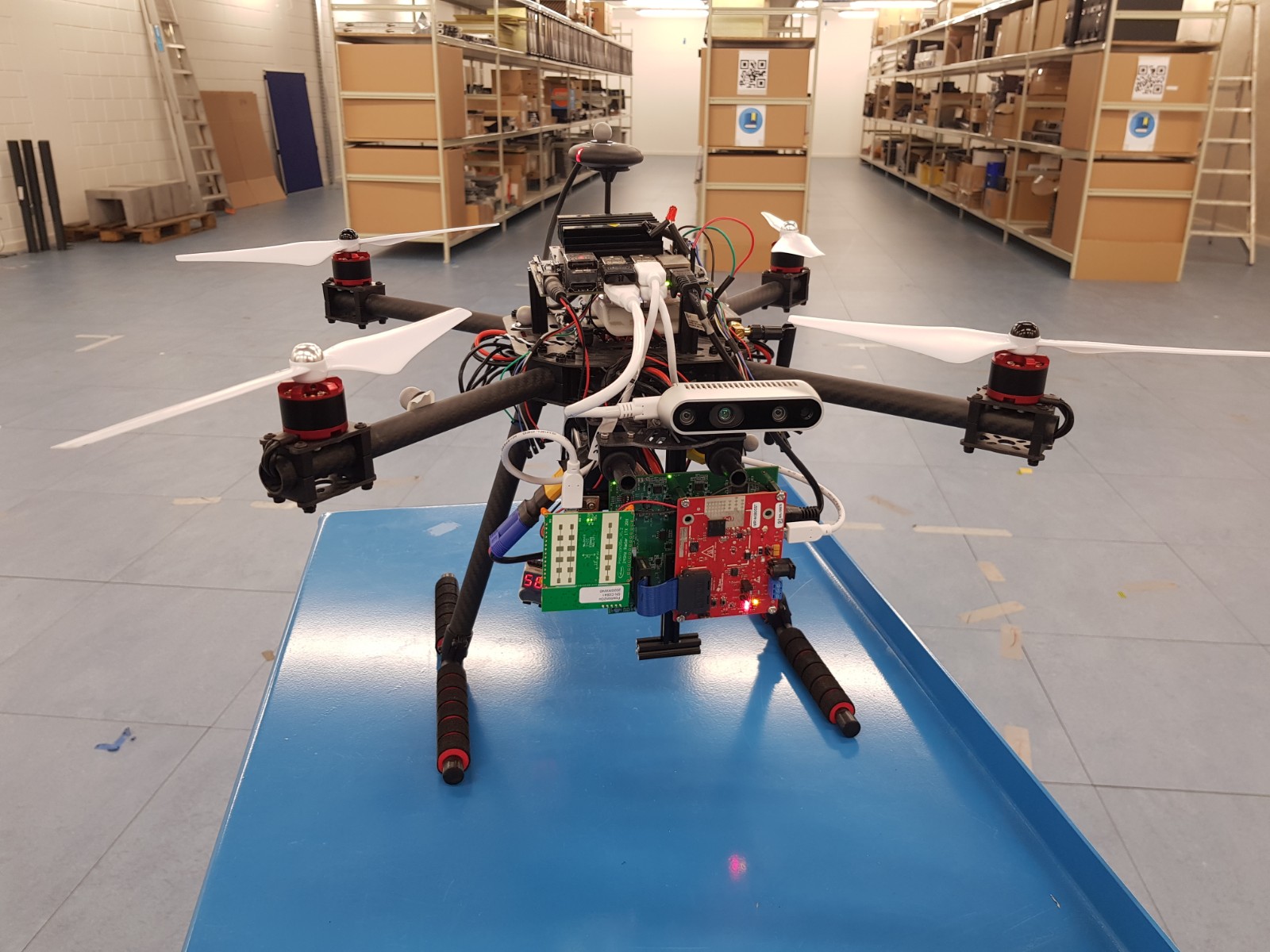}}
\caption{\textbf{Our NXP \textit{HoverGames} drone} with a 79 GHz Texas Instruments radar and an on-board camera.}
\label{figdr}
\end{figure}

Fig. \ref{figdr} shows the NXP \textit{HoverGames} drone that we used to acquire the radar data. This drone setup is well-suited for indoor tasks such as automated logistics, as it has sufficient power to carry payloads in the kilogram range. In addition, the off-the-shelve 79 GHz radar sensor is well-suited for precise obstacle ranging due to its centimetre-scale range resolution. The radar parameters are given in table \ref{radarparam}, where $L$ is the number of ADC samples per chirp, $T_f$ is the ADC sampling period, $N_c$ is the number of chirps in one frame, $T_c$ is the chirp period, $\alpha$ is the chirp slope and $d_{max}$ is the maximal distance that can be sensed.
\\

\begin{table}[htbp]
\begin{center}
\begin{tabular}{|c|c|c|c|c|c|}
\hline
$L$ & $T_f$  & $N_c$ &  $T_c$ & $\alpha$ & $d_{max}$\\
\hline
 256 & 160 ns & 128 & 80 $\mu$s & 50 MHz/$\mu$s & $18.3$ m\\
\hline
\end{tabular}
\caption{Radar parameters}
\label{radarparam}
\end{center}
\end{table}

\begin{figure}[htbp]
\centerline{\includegraphics[scale=0.4]{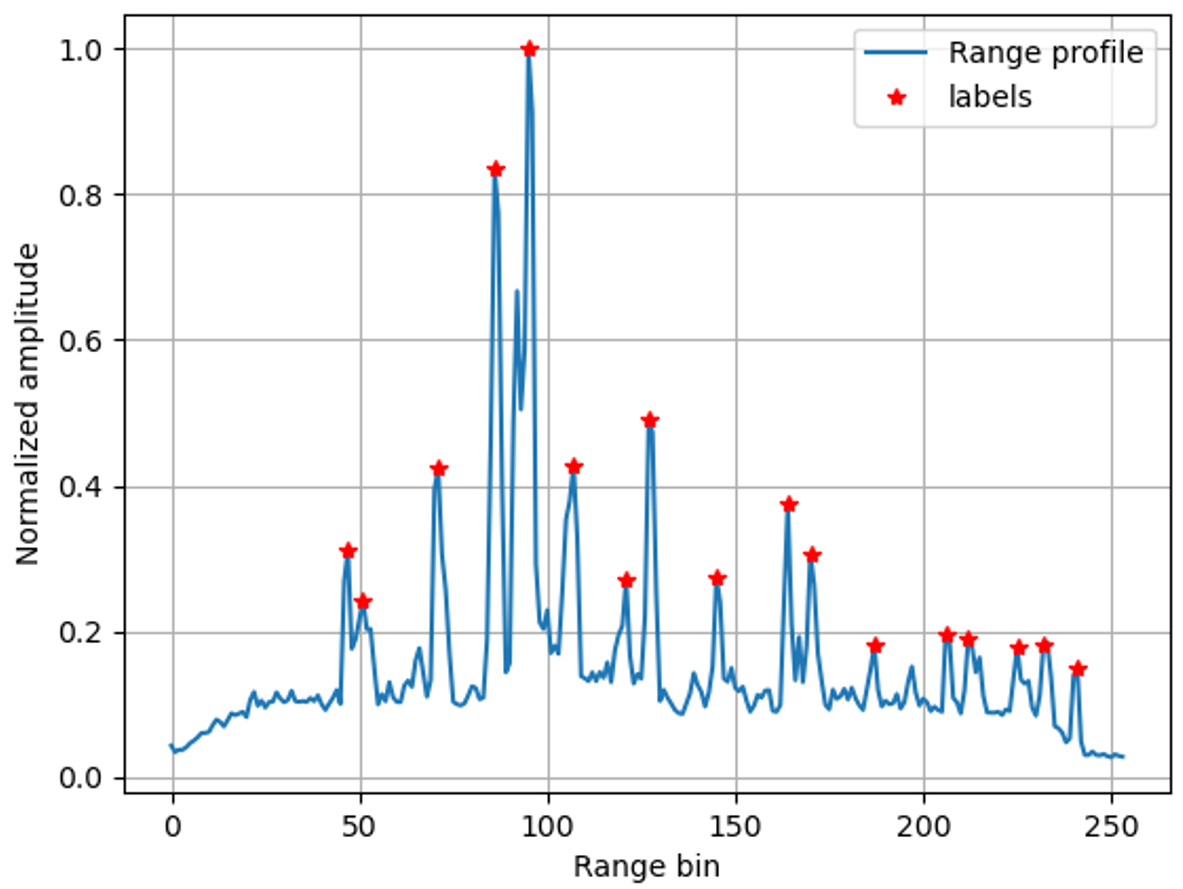}}
\caption{\textbf{An illustrative example RP} obtained when the drone was moving towards the shelves. }
\label{exlab}
\end{figure}

\begin{figure}[htbp]
\centerline{\includegraphics[scale=0.59]{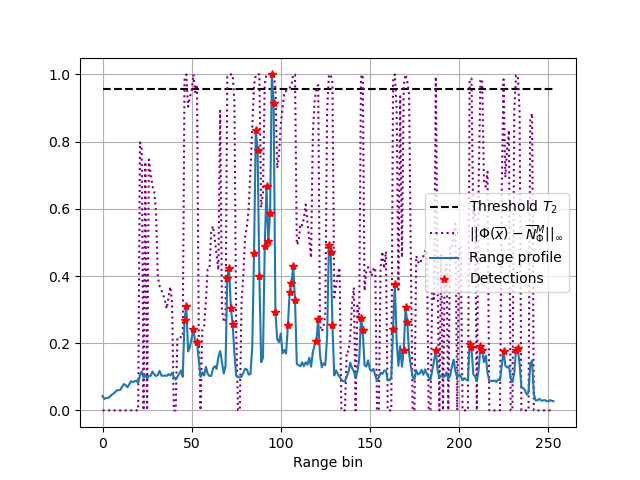}}
\caption{\textbf{Effect of applying our detector to the RP of Fig. \ref{exlab}.} The detection is performed as in (\ref{finfin}) by thresholding $||\Phi(\Bar{x}) - \Bar{N}_{\Phi}^m||_{\infty}$ (purple plot) against $T_2$. The detections are shown in red. The $D$ parameter was set to $15$ in this example.}
\label{shol}
\end{figure}

We navigated the drone in an indoor environment composed of benches, chairs, shelves and so on, and we obtained a dataset of nearly $400$ radar frames. Out of that dataset, we selected a subset of $100$ frames corresponding to the cases where the drone was flying towards a dense part of the environment, either towards the three shelves in Fig. \ref{figdr} or towards benches and chairs. Those cases represent the challenging, dense scenarios where missed detection can be detrimental to the drone. We finally obtained a dataset of 100 RPs and we visually labelled as a detection each significant peak in each of the 100 RPs. 
Fig. \ref{exlab} shows an illustrative example of a RP acquired when the drone was flying towards the shelves. The red stars denote the labelled indexes against which the detection algorithms will be compared. Fig. \ref{shol} shows the effect of applying our detector to the RP of Fig. \ref{exlab}, as per (\ref{finfin}), $||\Phi(\Bar{x}) - \Bar{N}_{\Phi}^m||_{\infty}$ is compared against a threshold level $T_2=0.95$ to obtain the subset of detected range bins. Fig. \ref{shol} shows the processing gain that our detector provides by mapping not only the strong RP peaks, but also the minute but significant peaks of the RP to high values of $||\Phi(\Bar{x}) - \Bar{N}_{\Phi}^m||_{\infty}$ (purple plot), while keeping this value lower for noise bins. 


\subsection{ROC-based Results}
\label{assess1}
\begin{figure}[htbp]
\centerline{\includegraphics[scale=0.55]{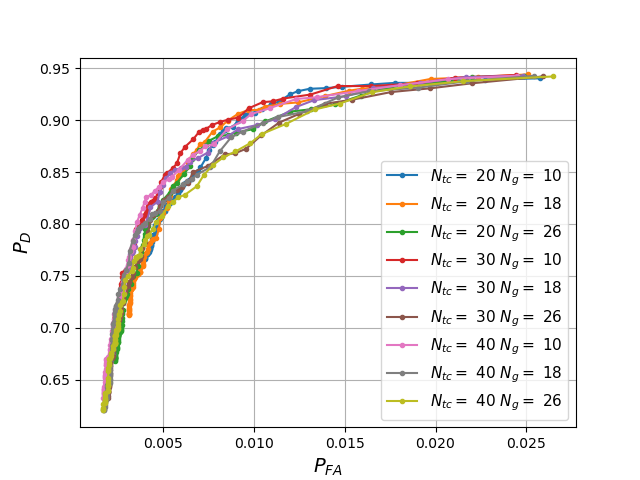}}
\caption{\textbf{ROC curves for our proposed detector}, for different combinations of $N_{tc}$ and $N_g$. $D = 15$.}
\label{ouroc}
\end{figure}

\begin{figure}[htbp]
\centerline{\includegraphics[scale=0.55]{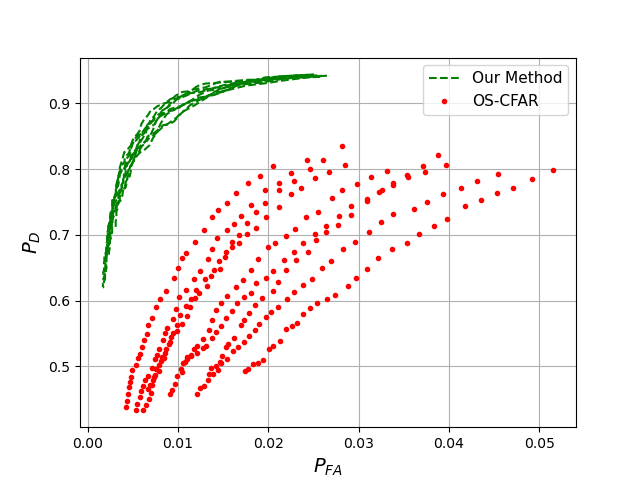}}
\caption{\textbf{k = 0.5 for OS-CFAR}. Average gain of 40\% on $P_D$ compared to OS-CFAR for $P_{FA} < 1\%$.}
\label{start}
\end{figure}

\begin{figure}[htbp]
\centerline{\includegraphics[scale=0.55]{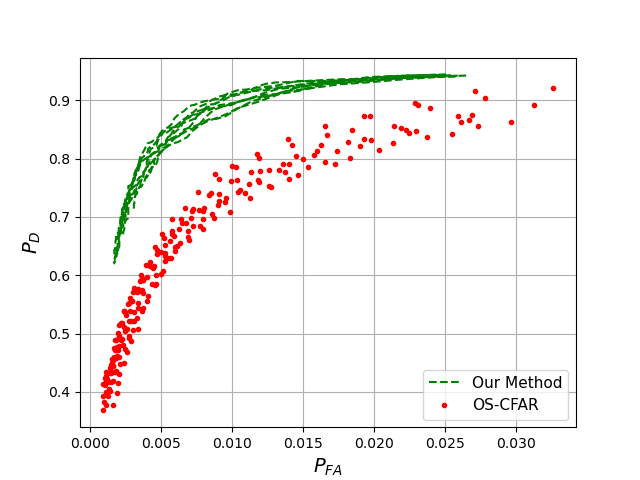}}
\caption{\textbf{k = 0.7 for OS-CFAR}. Average gain of 19\% on $P_D$ compared to OS-CFAR for $P_{FA} < 1\%$.}
\label{middle}
\end{figure}

\begin{figure}[htbp]
\centerline{\includegraphics[scale=0.55]{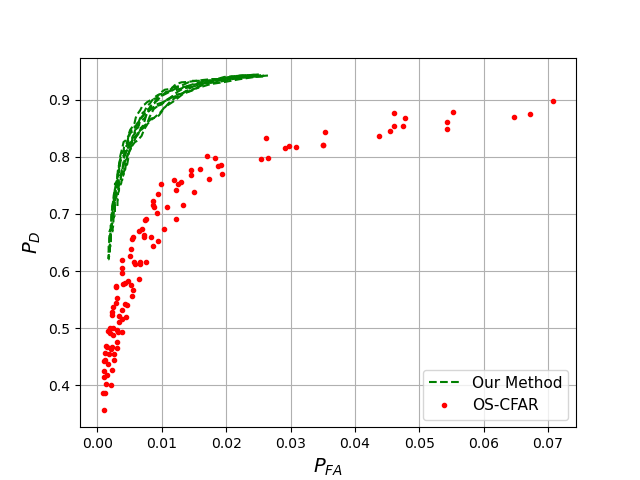}}
\caption{\textbf{k = 0.9 for OS-CFAR}. Average gain of 21\% on $P_D$ compared to OS-CFAR for $P_{FA} < 1\%$.}
\label{end}
\end{figure}

The standard method used to compare detectors is the generation of the ROC curves giving the probability of detection $P_{D}$ versus the probability of false alarm $P_{FA}$. $P_{D}$ is defined as the number of ground-truth detection labels found by the detector under test over the total number of ground-truth detection labels. Similarly, $P_{FA}$ is defined as the number of detections returned by the detector under test and not associated to any ground-truth detection label, over the total number of range-bin $L$. The higher the $P_D$ for a given $P_{FA}$, the better. We compute $P_D$ and $P_{FA}$ for both our method and OS-CFAR by applying them to each of the 100 RPs and by comparing their output to the labels. We consider that a detection is a false alarm when it lies at a distance larger than $5$ range bins ($35$ cm) from any label and \textit{vice versa} for a correct detection. We therefore obtain the number of detected labels $N_{Di}$ and the number of false alarms $N_{FAi}$ for each RP $i=1,...,100$. $P_D$ is then obtained as $\frac{1}{100}\sum_i (N_{Di}/l_i)$ where $l_i$ is the number of labelled indexes for RP $i$ while $P_{FA}$ is obtained as $\frac{1}{100}\sum_i (N_{FAi}/L)$ where $L=256$ is the number of range bins in the RPs. The tuples $(P_D,P_{FA})$ are computed for different threshold values $T_2$ for our detector (see section \ref{sec2}), different $\alpha$ and $k$ parameters for OS-CFAR (see section \ref{secchal}) and different numbers of training and guard cells ($N_{tc}$, $N_g$) for both methods. Each threshold value leads to a different tuple $(P_D,P_{FA})$ and the set of tuples is plotted as the ROC curve. For our proposed detector, we set the dimensionality $D$ of the transformation $\Phi(\xi)$ to 15 as it gave us the best results. Fig. \ref{ouroc} shows the ROC curves for our proposed detector. Fig. \ref{start} to \ref{end} jointly show the ROC curves for our proposed detector and for OS-CFAR, for different values of order parameter $k$ and for the same combinations of $N_{tc}$ and $N_g$ as in Fig. \ref{ouroc}. It is visually clear that our method significantly outperforms OS-CFAR without the requirement for computationally expensive \textit{sorting} operations per CUT. Our proposed algorithm always achieves a significantly larger $P_D$ for a given $P_{FA}$, for the different combinations of $k$, $N_{tc}$ and $N_g$, with a measured average gain of more than $\mathbf{19\%}$ between ROCs of the same $N_{tc}$ and $N_g$ on the achievable $P_D$ for a given $P_{FA}$, for $P_{FA}$ values smaller than 1\% (typical values used in practice).

\begin{figure}[htbp]
\centerline{\includegraphics[scale=0.55]{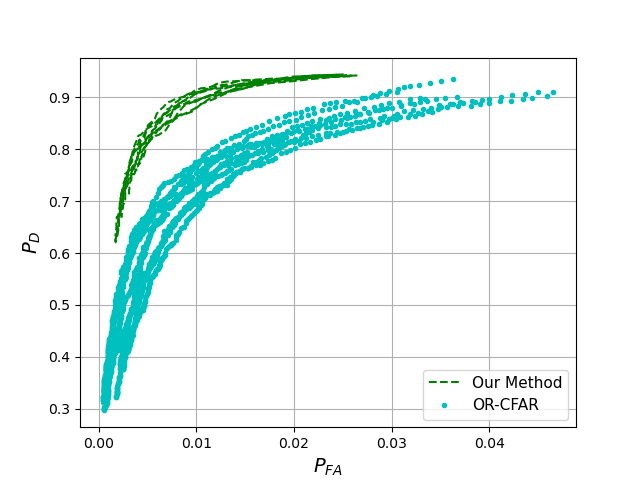}}
\caption{\textbf{OR-CFAR} with $\mathbf{\gamma=1.3}$. Average gain of 20\% on $P_D$ compared to CHA-CFAR for $P_{FA} < 1\%$.} 
\label{orcfar}
\end{figure}

\begin{figure}[htbp]
\centerline{\includegraphics[scale=0.55]{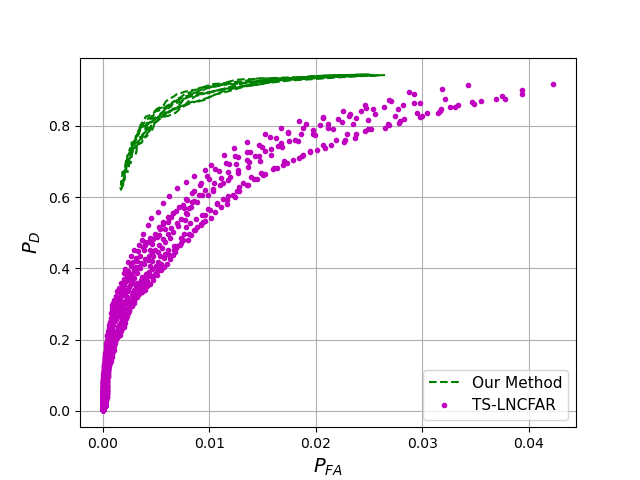}}
\caption{\textbf{TS-LNCFAR} with $\mathbf{\gamma=1.8}$. Average gain of 35\% on $P_D$ compared to CHA-CFAR for $P_{FA} < 1\%$.} 
\label{lscfar}
\end{figure}

\begin{figure}[htbp]
\centerline{\includegraphics[scale=0.55]{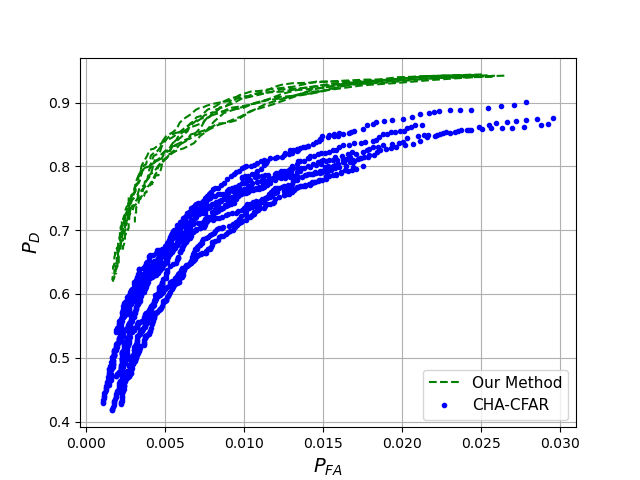}}
\caption{\textbf{CHA-CFAR} with $\mathbf{m=0.65\times N_{tc}}$. Average gain of 16\% on $P_D$ compared to CHA-CFAR for $P_{FA} < 1\%$.}
\label{chacfar}
\end{figure}

In addition, we compare our proposed detector to OR-CFAR, TS-LNCFAR and CHA-CFAR (Fig. \ref{orcfar}, \ref{lscfar} and \ref{chacfar}) by tuning the truncation parameter $\gamma$ of OR-CFAR and TS-LNCFAR, and the order parameter $m$ of CHA-CFAR such that each competing detector reaches its best performance on the indoor drone dataset used for the assessment. A trend similar to the OS-CFAR case is observed where our proposed detector systematically leads to a higher $P_D$ for a given $P_{FA}$ for our particular indoor scenario (OR-CFAR, TS-LNCFAR and CHA-CFAR are rather tailored for remote earth observation with higher-resolution SAR images). This shows the effectiveness of our new detection algorithm when used in indoor, industrial settings.

\subsection{Result confirmation with ground-truth data}
\label{conf}
\begin{figure}[htbp]
\centerline{\includegraphics[scale=0.32]{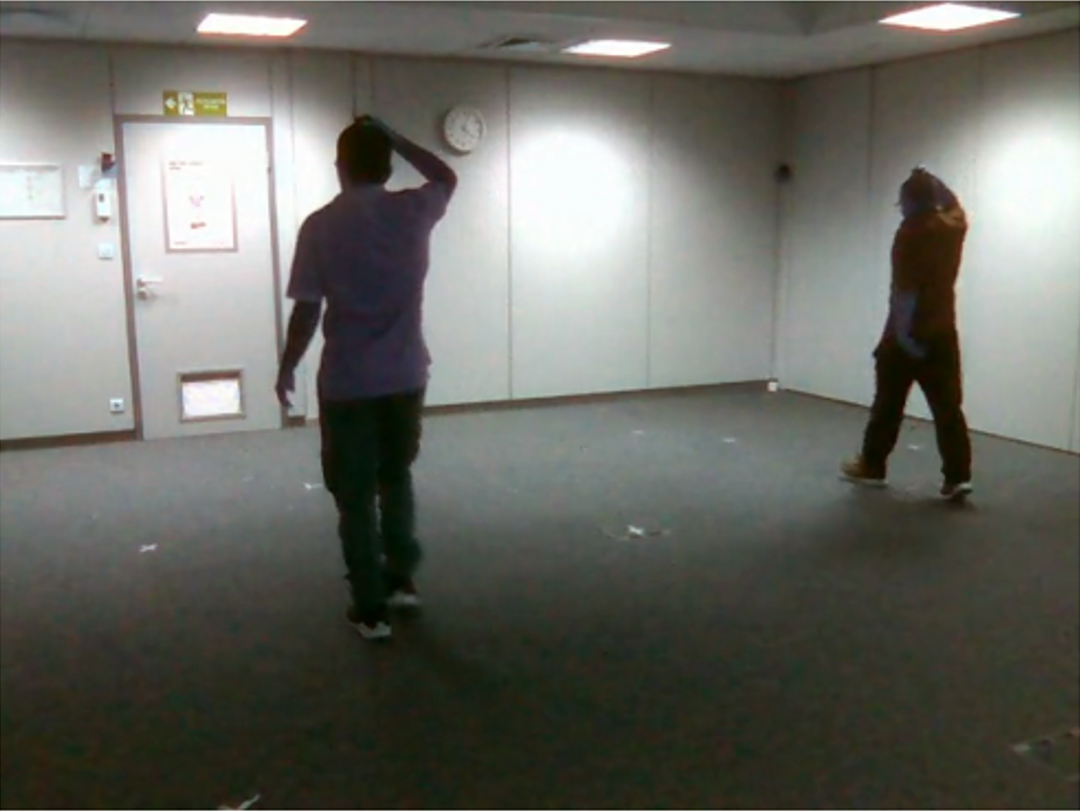}}
\caption{\textbf{Ground-truth and radar acquisition.} The two walking people hold their indoor GPS beacon on their head during the radar acquisition.}
\label{glimpse}
\end{figure}

As our goal is to detect as many targets as possible for a safe obstacle avoidance, the assessment presented in section \ref{assess1} has been conducted with manually labelled RP data, where each significant peak is marked as a potential obstacle. The potential downside of this visual labelling methodology is that noise and clutter could also lead to peaks in the RP, which, once labelled, would jeopardise the ROC assessment. On the other hand, the scenes acquired by our drone and radar are rich of obstacles. This motivates the fact that the overwhelming majority of labelled peaks would correspond to actual targets. Still, it is important to confirm this fact using a well-controlled, yet realistic radar dataset with ground-truth measurements. Therefore, we acquired a second RP dataset in an empty room by keeping our drone and radar setup static, with two people walking in front of it. During the radar acquisition, a \textit{Marvermind indoor GPS} device was used to acquire the ground-truth location of the two walking people. This scenario was chosen as it is more realistic than using usual corner reflector beacons (e.g., the scenario simulates a drone waiting for instructions with people walking in its neighbourhood). Yet, the chosen scenario enables precise ground-truth measurements using the indoor GPS. Fig. \ref{glimpse} depicts the acquisition scenario. 

\begin{figure}[htbp]
\centerline{\includegraphics[scale=0.58]{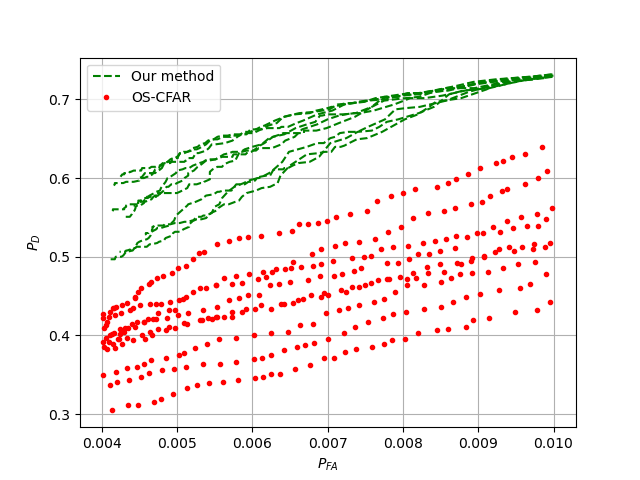}}
\caption{\textbf{k = 0.7 for OS-CFAR}. Average gain of 19 \% on $P_D$ compared to OS-CFAR for $P_{FA}$ values smaller than 1\%.}
\label{compara}
\end{figure}

Similar to section \ref{assess1}, a ROC-based comparison of our method against OS-CFAR has been conducted. The results are shown in Fig. \ref{compara}. The OS-CFAR parameter $k$ is set to $0.7$ as this value corresponds to the smallest gain between our proposed method and OS-CFAR in the previous assessment of section \ref{assess1}, Fig. \ref{middle}. The parameter $D$ of our proposed algorithm is set to $4$ in order to reach a desired $P_{fa}$ range of $<1\%$ (the impact of the $D$ parameter will be further detailed in section \ref{effd}). Compared to the results of section \ref{assess1}, Fig. \ref{compara} shows a similar trend where our proposed detector provides again a significant gain ($44\%$) on $P_D$ compared to OS-CFAR for the different combinations of $N_{tc}$ and $N_{g}$ (same as in Fig. \ref{ouroc}).

\begin{figure}[htbp]
\centerline{\includegraphics[scale=0.58]{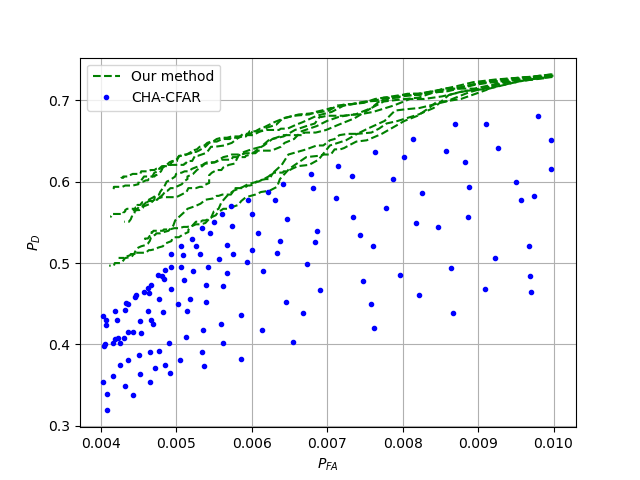}}
\caption{\textbf{$\mathbf{m = 0.4 \times N_{tc}}$ for CHA-CFAR}. CHA-CFAR achieves a better performance compared to OS-CFAR (Fig. \ref{compara}) but is still outperformed by our proposed method by an average gain of $14 \%$.} 
\label{compara2}
\end{figure}

In addition, Fig. \ref{compara2} shows the performance of CHA-CFAR against our proposed detector (OR-CFAR and TS-LNCFAR do not reach a $P_{D} > 10 \%$ on the tested $P_{FA}$ range and are therefore not shown). The order parameter $m$ of CHA-CFAR is chosen such that the detector performs at its best. Even though CHA-CFAR outperforms OS-CFAR, our proposed method still outperforms CHA-CFAR with an average gain of $32 \%$ between the corresponding curves. Again, this validates the effectiveness of our novel method for indoor radar sensing.

\subsection{Discussion}
\label{discusssec}
\begin{figure}[htbp]
\centerline{\includegraphics[scale=0.58]{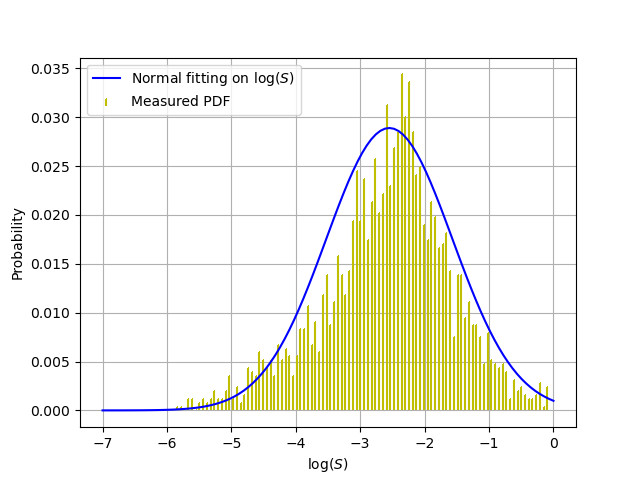}}
\caption{\textbf{Log-normal fitting.} The KS distance is $0.048$.}
\label{logno}
\end{figure}

\begin{figure}[htbp]
\centerline{\includegraphics[scale=0.58]{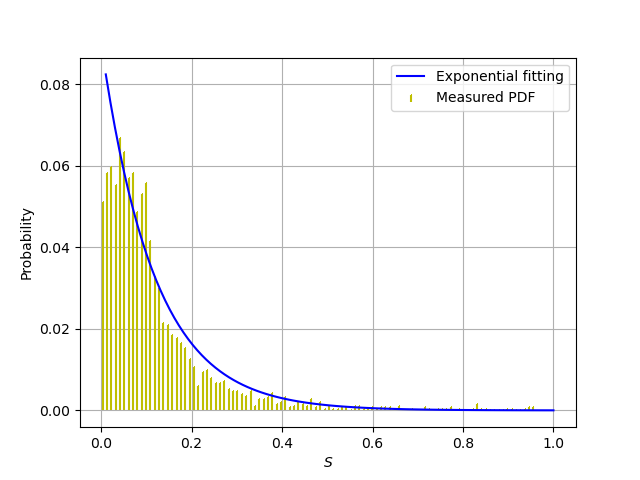}}
\caption{\textbf{Exponential fitting.} The KS distance is $0.056$.}
\label{expo}
\end{figure}

At this point, we have \textit{experimentally} shown that our proposed detector outperforms OS-CFAR, OR-CFAR, TS-LSCFAR and CHA-CFAR in our specific indoor sensing scenario. The reason for this is twofold. First, conventional CFAR methods and their state-of-the-art derivatives (see sections \ref{secchal} and \ref{sotalol}) perform detection by ML estimation of the clutter statistics given a prior clutter distribution model \cite{orcfar, lncfar}. Therefore, the performance of those conventional techniques degrade when the clutter distribution severely deviates from their prior hypothesis. In outdoor scenarios such as remote earth sensing, clutter distribution is often analytically well modelled by distributions such as Gaussian, log-normal and so on \cite{itshard} as the wave propagation paths are simpler. In contrast, indoor scenarios lead to severe multi-path effects as a large number of reflectors are present. In addition, floor reflectively is not always uniform which can lead to an increased number of clutter edges. Finally, compared to satellite SAR radars (as used in \cite{orcfar, lncfar, chacfar,gammacfar}), our radar setup provides a much more modest resolution which may lead to residual correlations between the bins in the RP, violating the assumption of inter-bin statistical independence \cite{book}. Fig. \ref{logno} and \ref{expo} show the \textit{goodness-of-fit} of an RP data distribution against the log-normal and the exponential models, which gave the smallest \textit{Kolmogorov-Smirnov} (KS) distances out of a set of typical hypothesis distributions (log-normal, Gaussian, Exponential, Gamma and Weibull) \cite{book}. For both hypothesis PDFs, the KS distance is above one order of magnitude larger than what is considered to be acceptable in ML-based CFAR estimation \cite{orcfar,chacfar}. 

The second reason behind the superiority of our method is the following. Conventional CFAR methods based on the OS principle (OS-CFAR and CHA-CFAR in this work) reject the outliers within the training cells by sorting the cells and choosing either the $n^{th}$ entry as the clutter statistics (as done in OS-CFAR) or either a combination of all entries above $n$ (as done in CHA-CFAR). For OS-CFAR, this means that a maximum of $N_{tc}-n$ outliers can be rejected, making it difficult to find a suitable $n$ in the dense, \textit{indoor} case (the number of outliers per training window is non-homogenous). CHA-CFAR performs slightly better than OS-CFAR in the indoor case (see Fig. \ref{compara} vs. Fig \ref{compara2}) as it estimates the clutter statistics through a combination of the training cell entries above $n$. This \textit{harmonic averaging} combination is such that entries with large values have a smaller contribution than entries with small values. Consequently, the order $n$ can be chosen smaller than in the OS-CFAR case, which allows an adaptive rejection of an increased number of outliers in each training window (in contrast to the \textit{hard} rejection of OS-CFAR).

On the other hand, modern hyperspectral detection techniques such as SVDD or KRX \cite{nasr} (from which our proposed method is derived) do not estimate the clutter statistics \textit{explicitly} and therefore, must not rely on a prior PDF model for ML estimation. Rather, those techniques solve a binary classification problem by expressing the local data (training cells and CUT) in a high-dimensional feature space $\mathcal{F}$ where the separability between samples is enhanced \cite{nasr}. Simple decision contours in $\mathcal{F}$ will then correspond to much more complex decision boundaries in the original space, enabling high-performance detection in unknown and complex clutter statistics \cite{nasr}. As experimentally shown in this paper, such a detection principle is especially attractive for \textit{indoor} radar sensing where the \textit{goodness-of-fit} of usual PDF models is not as good as in the \textit{outdoor} case due to the complexity of the sensed environment. Finally, it must be noted that such high-dimensional, non-linear methods come at the cost that finding analytical expressions for e.g., setting the detection threshold $T_2$ to meet a target $P_{FA}$ is mathematically intractable. This is not only the case for our proposed detector (which uses (\ref{finalphi}) as a discontinuous non-linearity function), but is also the case for the original KRX formulation (which uses a smooth RBF kernel \cite{kerrx}).

\subsection{Effect of the $D$ parameter}
\label{effd}
\begin{figure}[htbp]
\centerline{\includegraphics[scale=0.58]{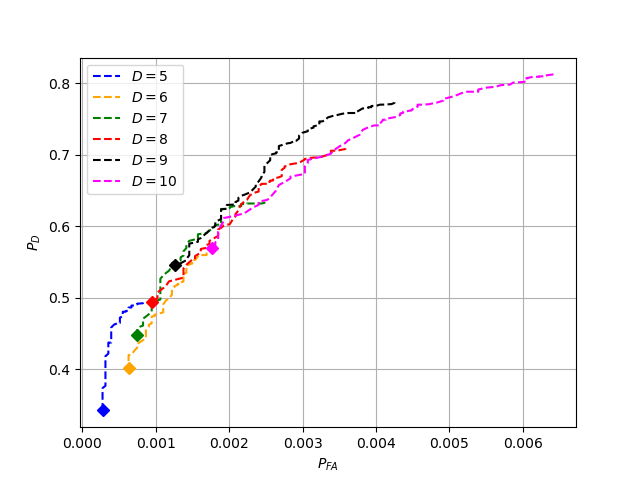}}
\caption{\textbf{ROC curves for different $D$ values.} The diamond-shaped points denote the $(P_D, P_{FA})$ achieved in the limit of $T_2 \rightarrow 1$.}
\label{effectd}
\end{figure}

In order to evaluate the effect of the $D$ parameter (dimensionality of the transformation $\Phi(\xi)$ in (\ref{finalphi})), we fix $N_{tc}=20$, $N_g=10$ and we evaluate the ROC curves of the proposed detector using the same dataset used in section \ref{assess1} for a range of $D$ values. The results are shown in Fig. \ref{effectd}.

As Fig. \ref{effectd} shows, the $D$ parameter affects the minimum reachable $P_{FA}$. This can be explained as follows. Each element of the vector $\Bar{N}_{\Phi}^m$ in (\ref{bad1}) can take a maximum value of $1$ because of the normalization by $\Gamma$ during the \textit{centroid} computation and following the definition of $\Phi(\xi)$ in (\ref{finalphi}) where each element of the vector $\Phi(\xi)$ is either $0$ or $1$. Indeed, if all elements of $\Phi(X_{tc}[j]), \forall j$ are equal to $1$ (which is the limit case of the $\Phi(\xi)$ encoding), the evaluation of (\ref{bad1}) will result in a vector $\Bar{N}_{\Phi}^m$ with all of its components equal to $1$ because of the normalization by $\Gamma$. From (\ref{finfin}), we can therefore write:  
\begin{equation}
    ||\Phi(\Bar{x}) - \Bar{N}_{\Phi}^m||_{\infty} \leq 1
    \label{wecanwrite}
\end{equation}
as the entries of $\Phi(\Bar{x})$ are either $0$ or $1$ by the definition of $\Phi(\xi)$ in (\ref{finalphi}). Therefore, only values of $T_2$ in (\ref{finfin}) below $1$ will affect the $P_D$ and $P_{FA}$ of the proposed detector. The $P_{FA}$ achieved in the limit of $T_2 \rightarrow 1$ for a certain $D$ value will thus be the minimal reachable $P_{FA}$ for that $D$. Then, the smaller $D$ is, the less quantization slots are available during the binary coding induced by $\Phi(\xi)$ in (\ref{finalphi}). As a consequence, the likelihood that the training cells in the vector $X_{tc}$ have a $1$ in the same quantization slot ($i$ in (\ref{finalphi})) will grow, leading to larger elements in $\Bar{N}_{\Phi}^m$, which will lower the evaluation of $||\Phi(\Bar{x}) - \Bar{N}_{\Phi}^m||_{\infty}$, lowering both the $P_D$ and the $P_{FA}$. This can clearly be observed in Fig. \ref{effectd}, where the smaller the value of $D$, the smaller the minimum reachable $P_{FA}$ is. Naturally, a similar discussion holds concerning the maximum reachable $P_D$ in the limit of $T_2 \rightarrow 0$, where, the smaller the $D$, the smaller the maximum reachable $P_D$ will be.

\subsection{Ablation Studies}
\label{abl}
\begin{figure}[htbp]
\centerline{\includegraphics[scale=0.58]{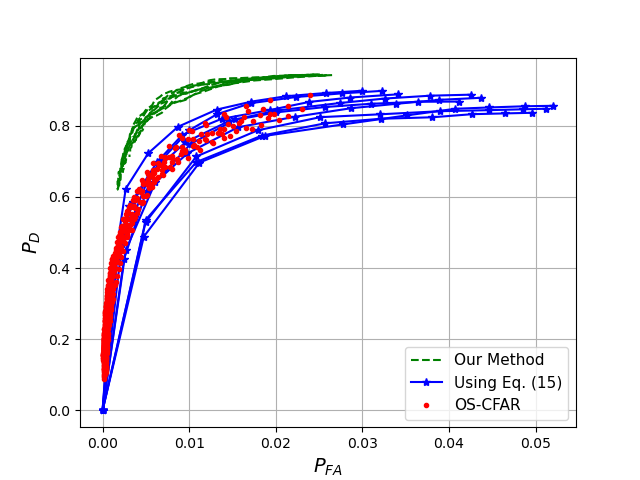}}
\caption{\textbf{ROC curves for different methods} (k = 0.7 for OS-CFAR). Our method uses (\ref{bad1}).}
 \label{ab1}
\end{figure}

It is now important to verify the impact of the choices made for our detector in section \ref{sec2} on the performance of the algorithm. We have introduced a heuristic rule (\ref{bad1}) that replaced the usual computation of (\ref{eqlastt}) by a \textit{centroid} since we did not expect the projection of the RPs in the feature space $\mathcal{F}$ to be Gaussian, as opposed to the assumption of (\ref{eqlastt}). Fig. \ref{ab1} compares our heuristic rule against (\ref{eqlastt}) and against OS-CFAR with $k=0.7$ (as it was the closest to our method in terms of performance). As expected, using (\ref{eqlastt}) results in a significant loss of performance compared to (\ref{bad1}). This experimentally validates the need for our heuristic rule (\ref{bad1}). 

\begin{figure}[htbp]
 \centerline{\includegraphics[scale=0.58]{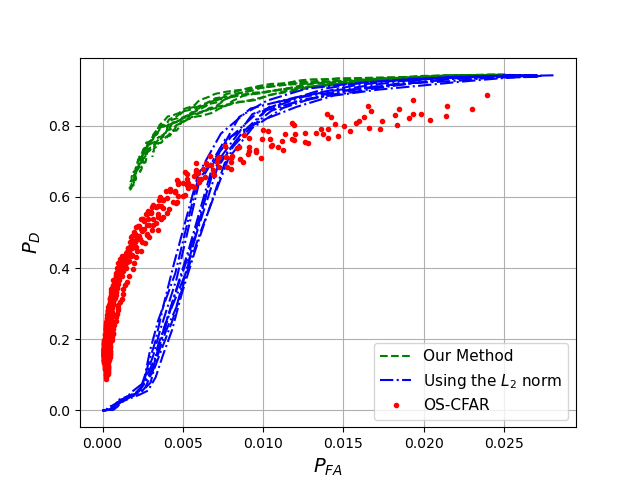}}
 \caption{\textbf{ROC curves for different methods} (k = 0.7 for OS-CFAR). Our method uses the $L_{\infty}$ norm.}
 \label{ab2}
 \end{figure}

Finally, we have decided to use the $L_{\infty}$ norm in (\ref{finfin}) instead of the $L_2$ norm and we gave motivations for this choice is section \ref{sec2}. Fig. \ref{ab2} compares the use of the $L_{\infty}$ norm against the $L_2$ norm. 
 It is interesting to see that OS-CFAR performs better for low $P_{FA}$ values compared to using the $L_{2}$ norm in (\ref{finfin}). Finally, it is visually clear that our method (using the $L_{\infty}$ norm) leads to a significantly better performance compared to the $L_2$ norm for $P_{FA} < 0.015$.

\subsection{System implementation and hardware metrics}
\label{hardmet}
We have implemented our proposed detector in an \textit{ARM Cortex R4F} processor running at 40 MHz and integrated in the 79 GHz radar chip used during the data acquisitions. Our detector returns a list of populated range bins which are then passed to a conventional beamformer algorithm \cite{review} to compute the angle of arrival of each detection. As OS-CFAR involves an expensive \textit{search operation} for each cell under test, it is already clear that our method is less computationally expensive. It is, however, more meaningful to compare our method against the standard detection API provided by the chip vendor, which uses a \textit{hardware-accelerated} CA-CFAR applied on the full RD map (2D detection performed along each range and Doppler bin) followed by additional clutter filtering. Compared to our detector written in plain C, their detector is not running in the CPU but in a dedicated hardware accelerator specially tailored for their 2D CA-CFAR  implementing vectored computation. This hardware acceleration is needed to meet the inter radar frame timing as, unlike our single-pass detection, they perform detection along each range and Doppler bin ($256 \times 128$ detection passes vs. 1 in our case). Then, as CA-CFAR is subject to a high amount of clutter false alarms, their API performs a final clutter filtering step and the detection list is returned to a conventional beamformer algorithm (identical to the one used with our proposed detector) to compute the angle of arrival of each detection. 

We have compare our method against theirs in term of \textit{average inter-frame processing time margin}, meaning the \textit{time slack} available between the execution of each complete radar frame processing (reception of raw ADC data - range-Doppler processing - detection - angle of arrival estimation). Table \ref{timemarg} shows the measured average inter-frame processing time margin.

\begin{table}[htbp]
\begin{center}
\begin{tabular}{|c|c|}
\hline
\textbf{Method} & \textbf{Time Slack [ms]}   \\
\hline
Standard  & 26.7 \\
\hline
\textbf{Ours} & \textbf{45.7} \\
\hline
\end{tabular}
\caption{\textbf{Measured average inter-frame processing time margin.} The radar frame rate is set to 10 FPS.}
\label{timemarg}
\end{center}
\end{table}

\begin{figure}[h!]
\centerline{\includegraphics[scale=0.7]{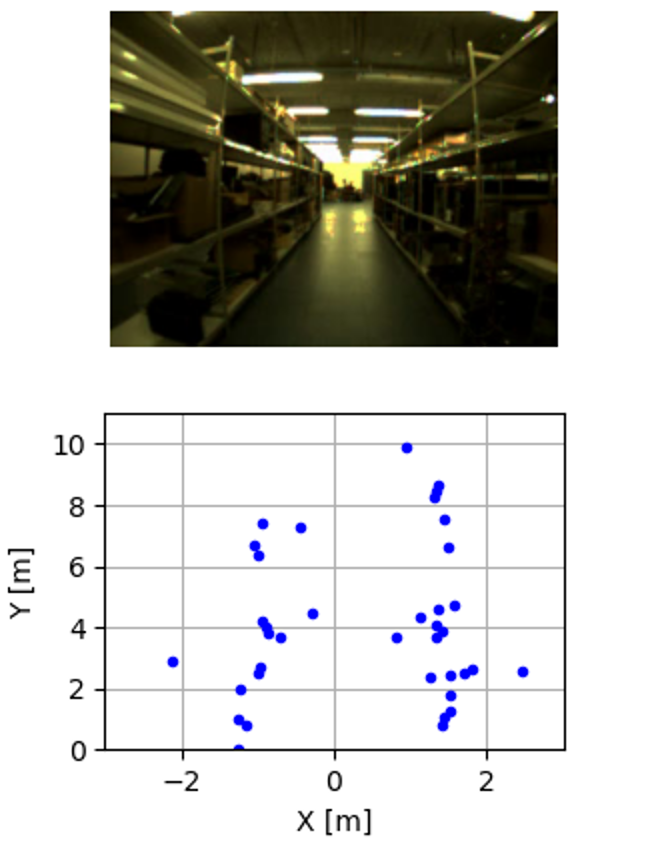}}
\caption{\textbf{Radar-camera snapshot of our drone flying through industrial aisles.} The right and left shelves can easily be identified.}
\label{addi}
\end{figure}

Table \ref{timemarg} clearly shows that our method executes nearly two times faster, providing a significant increase in the time slack available between the execution of each frame processing step, even though our method does not rely on any dedicated hardware acceleration. In addition, a video in our supplementary material showcases the reduction in measured CPU load reported by the graphical interface provided by the radar vendor. This additional time slack enables our system to run at frame rates $\sim$35 FPS before the dead-lock of the MCU (due to the fact that radar chirps are incoming at speeds larger than what the MCU can process), while the standard method can run at maximum 17 FPS before the dead-lock of the MCU. This larger frame rate allows our radar system to detect obstacles faster, enabling safer drone flights. Furthermore, this additional time slack can be used to execute more complex angle of arrival algorithms after the detection step to enhance the poor angle resolution of the conventional beamformer \cite{review}. Finally, Fig. \ref{addi} shows a radar-camera snapshot of our drone flying through industrial shelves with our proposed detector running on-line in the MCU of the radar chip at $30$FPS, with a conventional beamformer \cite{review} estimating the angle of arrival of the detections. The aisles can clearly be observed thanks to the high $P_D$ over $P_{FA}$ ratio that our detector provides, which qualitatively shows the effectiveness of our detector. A video showing the full flight can be found in the supplementary material.

\section{Conclusion}
\label{sec4}
This paper has focused on reliable radar target detection in dense indoor environments for drone applications requiring obstacle avoidance for safe navigation. We have proposed a novel radar target detection algorithm with low computational cost suitable for near-sensor implementation in edge devices. It is shown that the proposed detector outperforms OS-CFAR with more than 19\% higher $P_D$ and outperforms CHA-CFAR with more than 16\% higher $P_D$ for the same $P_{FA}$ in our indoor drone navigation scenario where missed detection can be detrimental, while being less computationally complex than both OS-CFAR and CHA-CFAR (as no sorting operations are needed). To the best of our knowledge, this work improves the state of the art for high-performance yet low-complexity radar detection in very dense indoor sensing applications.

\section*{Acknowledgment}
This work is partially supported by the Flanders AI research program. We would like to thank Dr. Aliakbar Gorji for acquiring the dataset used in section \ref{conf}.

\begin{IEEEbiography}[{\includegraphics[width=1in,height=1.5in,clip,keepaspectratio]{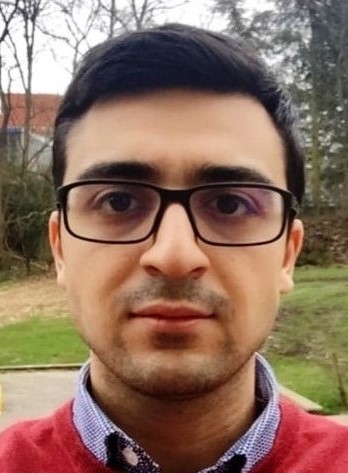}}]{Ali Safa}
(Student Member, IEEE) received the MSc degree (summa cum laude) in Electrical Engineering from the Universit\'e Libre de Bruxelles, Belgium. He joined IMEC and the Katholieke Universiteit Leuven (KU Leuven), Belgium in 2020 where he is currently working toward the PhD degree on AI-driven processing for extreme edge applications.  
\end{IEEEbiography}

\begin{IEEEbiography}[{\includegraphics[width=1in,height=1.3in,clip,keepaspectratio]{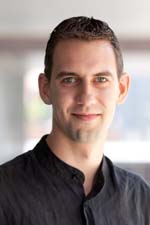}}]{Tim Verbelen} received his M.Sc. and Ph.D. degrees in Computer Science Engineering at Ghent University in 2009 and 2013 respectively. Since then, he is working as a senior researcher for Ghent University and imec. His main research interests include perception and control for autonomous systems using deep learning techniques and high-dimensional sensors such as camera, lidar and radar. In particular, he is active in the domain of representation learning and reinforcement learning, inspired by cognitive neuroscience theories such as active inference.

\end{IEEEbiography}

\begin{IEEEbiography}[{\includegraphics[width=1in,height=1.5in,clip,keepaspectratio]{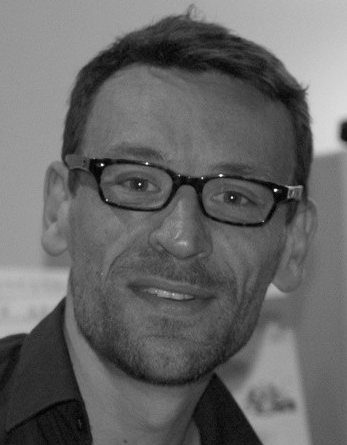}}]{Lars Keuninckx}
received a Master of Engineering in telecommunications from Hogeschool Gent in 1996 after which he worked in industry for several years, designing electronics for automotive, industrial and medical applications. He received a Bachelor in Physics in 2009 and a PhD in Engineering in 2016, both from the Vrije Universiteit Brussel. His interests include the applications of complex dynamics and reservoir computing. He joined imec in 2019 where he is involved in the design of neuromorphic systems.
\end{IEEEbiography}

\begin{IEEEbiography}[{\includegraphics[width=1in,height=10in,clip,keepaspectratio]{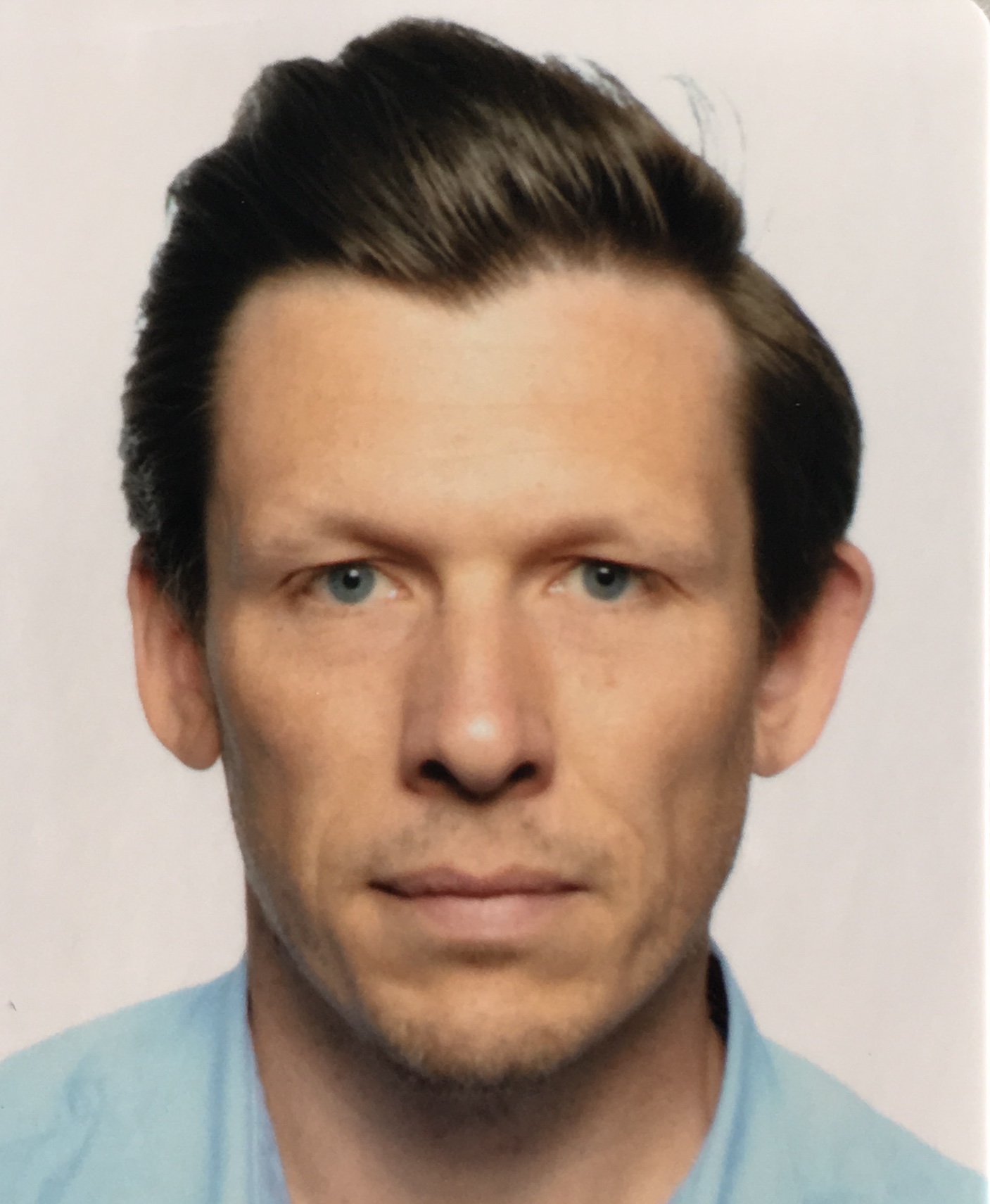}}]{Ilja Ocket}
(Member, IEEE) received the MSc and the PhD degrees in Electrical Engineering from KU Leuven, Leuven, Belgium, in 1998 and 2009, respectively. He currently serves as program manager for neuromorphic sensor fusion at the IoT department of imec, Leuven, Belgium. His research interests include all aspects of heterogeneous integration of highly miniaturized millimeter wave systems, spanning design, technology and metrology. He is also involved in research on using broadband impedance sensing and dielectrophoretic actuation for lab-on-chip applications.


\end{IEEEbiography}

\begin{IEEEbiography}[{\includegraphics[width=1in,height=10in,clip,keepaspectratio]{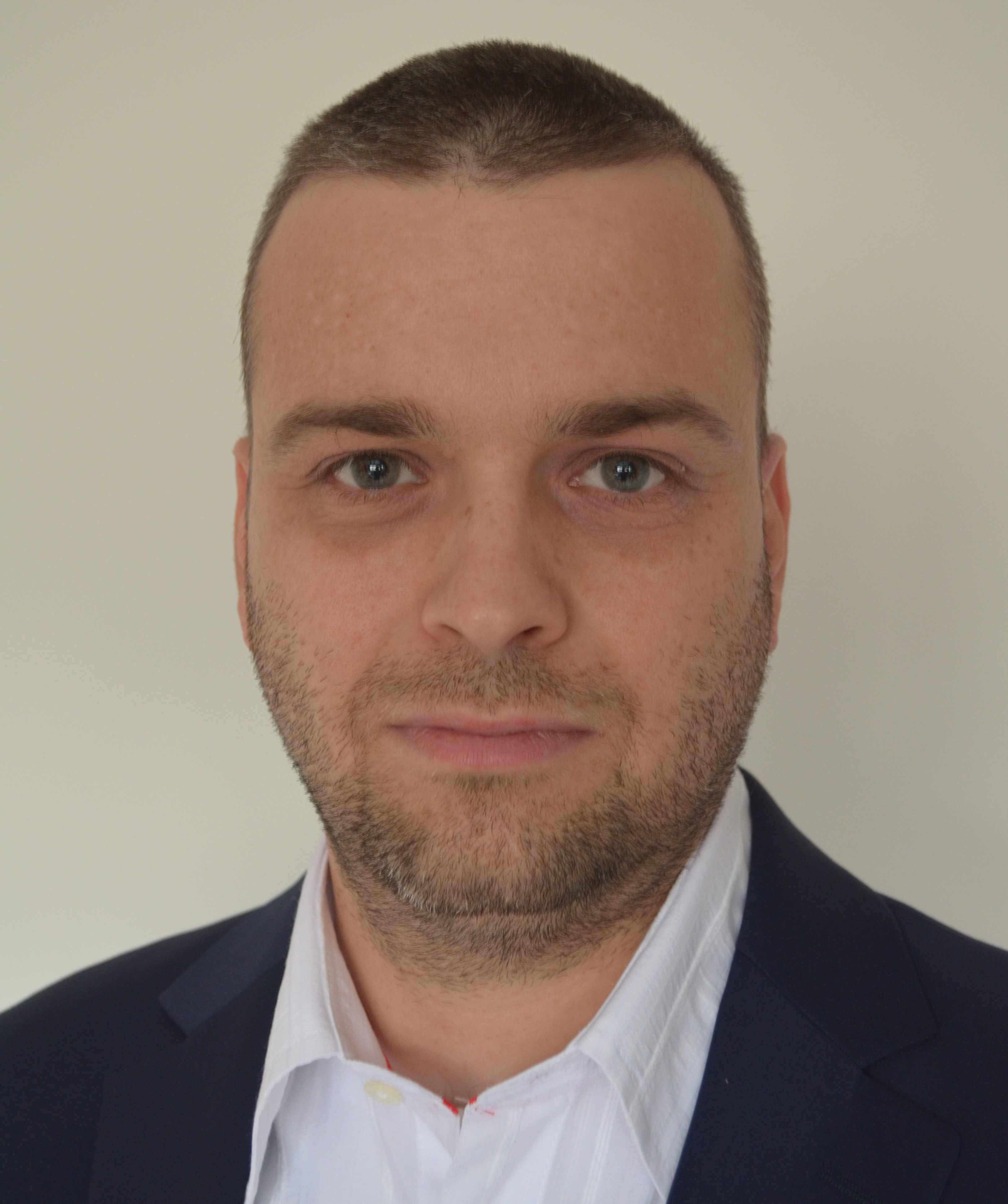}}]{Mathias Hartmann}
graduated from the Faculty of Electrical Engineering and Information Technology of the University of Technology Dresden. After obtaining a master’s degree in the field of Information and System Technology, he joined imec vzw (Leuven, Belgium) in 2008 working as a Research Engineer in the Architecture and Compiler team and afterwards as the technical lead for the PHADAR automotive radar program. Currently, he is engaged in the Neuromorphic Radar program.
\end{IEEEbiography}

\begin{IEEEbiography}[{\includegraphics[width=1in,height=10in,clip,keepaspectratio]{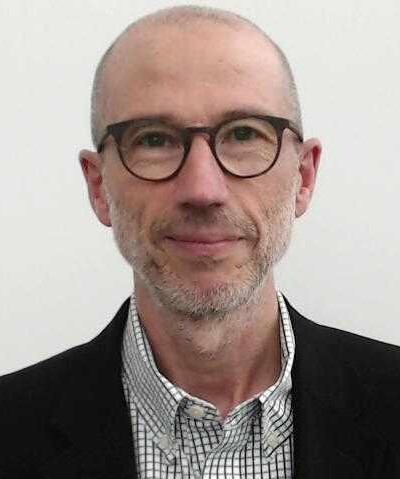}}]{Andr\'e Bourdoux}
(Senior Member, IEEE)
received the M.Sc. degree
in electrical engineering from the Universit\'e Catholique de Louvain-la-Neuve, Belgium,
in 1982. In 1998, he joined IMEC, where he is currently a Principal Member of Technical Staff with
the Internet-of-Things Research Group. He is a
system-level and signal processing expert for both
the mm-wave wireless communications and radar
teams. He has more than 15 years of research experience in radar systems and 15 years of research
experience in broadband wireless communications. He holds several patents
in these fields. He has authored or coauthored over 160 publications in books
and peer-reviewed journals and conferences. His research interests include
advanced signal processing, and machine learning for wireless physical layer
and high-resolution 3D/4D radars.
\end{IEEEbiography}

\begin{IEEEbiography}[{\includegraphics[width=1.1in,height=1.4in,clip,keepaspectratio]{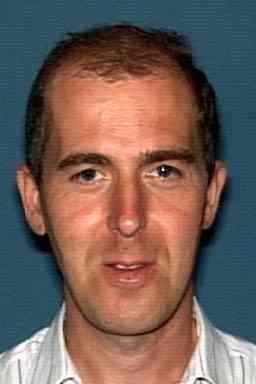}}]{Francky Catthoor}
(Fellow, IEEE) received  a Ph.D. in electrical engineering from KU Leuven,
Belgium in 1987.  Between 1987 and 2000, he has headed several research
domains in the area of synthesis techniques and architectural methodologies.
Since 2000 he is strongly involved in other activities at IMEC including
co-exploration of application, computer architecture and deep
submicron technology aspects, biomedical systems and IoT sensor nodes,
and photo-voltaic modules combined with renewable energy systems,
all at IMEC Leuven,  Belgium. Currently, he is an IMEC senior fellow.
He is also part-time full professor at the Electrical Engineering department of the KU Leuven (ESAT).
He has been associate editor for several IEEE and ACM journals and was elected IEEE fellow in 2005.

\end{IEEEbiography}

\begin{IEEEbiography}[{\includegraphics[width=1.1in,height=1.4in,clip,keepaspectratio]{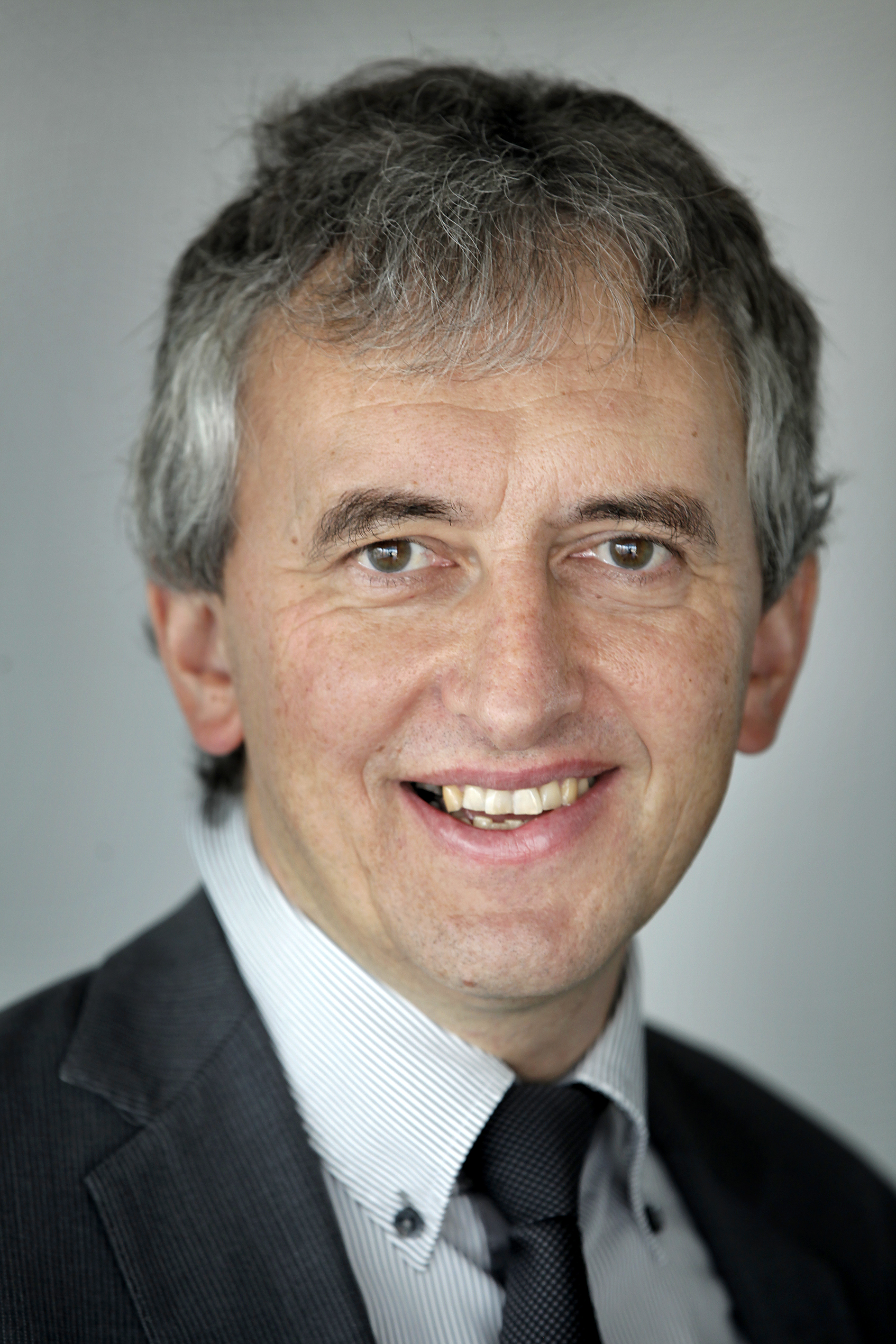}}]{Georges G.E Gielen}
(Fellow, IEEE) received the MSc and PhD degrees in Electrical Engineering from the Katholieke Universiteit Leuven (KU Leuven), Belgium, in 1986 and 1990, respectively. He currently is Full Professor in the MICAS research division at the Department of Electrical Engineering (ESAT) at KU Leuven. Since 2020 he is Chair of the Department of Electrical Engineering. His research interests are in the design of analog and mixed-signal integrated circuits, and especially in analog and mixed-signal CAD tools and design automation. He is a frequently invited speaker/lecturer and coordinator/partner of several (industrial) research projects in this area, including several European projects. He has (co-)authored 10 books and more than 600 papers in edited books, international journals and conference proceedings. He is a 1997 Laureate of the Belgian Royal Academy of Sciences, Literature and Arts in the discipline of Engineering. He is Fellow of the IEEE since 2002, and received the IEEE CAS Mac Van Valkenburg award in 2015 and the IEEE CAS Charles Desoer award in 2020. He is an elected member of the Academia Europaea.
\end{IEEEbiography}

\end{document}